\documentclass{article}
\usepackage[T1]{fontenc}
\usepackage{iclr2027_conference,times}
\iclrfinalcopy
\usepackage[hyphens]{url}
\usepackage{graphicx}
\usepackage{amsmath,amssymb,amsthm,mathtools}
\usepackage{enumitem}
\usepackage{algorithm}
\usepackage{algorithmic}
\usepackage{xcolor}
\usepackage{booktabs}
\usepackage{multirow}
\definecolor{darkblue}{RGB}{20,60,140}
\usepackage{hyperref}
\hypersetup{
  colorlinks=true,
  linkcolor={darkblue},
  citecolor={darkblue},
  urlcolor={darkblue},
  bookmarksnumbered=true,
  bookmarksopen=true,
  pdfstartview={FitH},
}

\graphicspath{{figs/}{figures/}}

\newtheorem{theorem}{Theorem}
\newtheorem{proposition}[theorem]{Proposition}
\newtheorem{lemma}[theorem]{Lemma}
\newtheorem{corollary}[theorem]{Corollary}
\newtheorem{definition}[theorem]{Definition}
\newtheorem{assumption}[theorem]{Assumption}

\DeclareMathOperator*{\E}{\mathbb{E}}
\newcommand{\F}{\mathcal{F}}
\newcommand{\D}{\mathcal{D}}

\newcommand{\G}{\mathcal{G}}
\newcommand{\Lt}{\mathcal{L}_{\mathrm{task}}}
\newcommand{\ow}{overthink-wrong}
\newcommand{\sco}{stay-correct}

\IfFileExists{numbers.tex}{\newcommand{\eZeroOverthinkDelta}{15.8}
\newcommand{\eZeroPeakT}{14}

\newcommand{\eZeroCompFinal}{85.3}

\newcommand{\eZeroSlopeMid}{1.04}
\newcommand{\eZeroRTwoMid}{0.82}

\newcommand{\eZeroFloorMatched}{0.05}
\newcommand{\eZeroMeanDev}{0.010}

\newcommand{\lateDriftAcc}{85.4}

\newcommand{\rsOWorig}{10}
\newcommand{\rsOWany}{3}
\newcommand{\rsSCorig}{40}
\newcommand{\rsSCstable}{38}
\newcommand{\rsN}{80}

\newcommand{\rsSCacc}{76.2}
\newcommand{\rsSCtokens}{15771}
\newcommand{\rsSharedR}{-0.21}
\newcommand{\aimeChains}{120}
\newcommand{\aimeOW}{0}
\newcommand{\aimeFinalAcc}{17.5}
\newcommand{\aimeCapRate}{88}
\newcommand{\aimeCurveStart}{0.9}
\newcommand{\aimeCurveEnd}{10.4}
\newcommand{\aimeCappedN}{106}
\newcommand{\fvPairs}{160}
\newcommand{\fvAgree}{81.6}
\newcommand{\fvCorrAgree}{85.7}
\newcommand{\fvNoAns}{69.4}
\newcommand{\gridFineOW}{10}
\newcommand{\gridCoarseOW}{9}
\newcommand{\gridFineOWsevB}{9}
\newcommand{\gridCoarseOWsevB}{3}

\newcommand{\wpRate}{3.2}
\newcommand{\wpCI}{[1.0\%,6.4\%]}
\newcommand{\wpUnw}{4.7}
\newcommand{\wpEnrich}{4.6}
\newcommand{\wpOWpop}{10}
\newcommand{\aimeElig}{12}
\newcommand{\aimeEligN}{120}
\newcommand{\aimeUpper}{25}

\newcommand{\rEightN}{2936}
\newcommand{\rEightOW}{100}
\newcommand{\rEightRate}{3.4}
\newcommand{\rEightCI}{[2.4\%,4.5\%]}
\newcommand{\rEightSC}{1989}
\newcommand{\rEightNC}{598}
\newcommand{\rEightLB}{249}
\newcommand{\rEightSibOW}{23.8}
\newcommand{\rEightSibSC}{2.1}

\newcommand{\vCommit}{0.284}
\newcommand{\vVerdict}{0.249}
\newcommand{\sevBvCommit}{0.221}
\newcommand{\sevBvVerdict}{0.244}
\newcommand{\vCommitCI}{[0.263,0.305]}
\newcommand{\sevBvCommitCI}{[0.200,0.245]}
\newcommand{\vGap}{0.036}
\newcommand{\sevBvGap}{-0.023}
\newcommand{\vGapCI}{[+0.007,+0.063]}
\newcommand{\sevBvGapCI}{[-0.050,+0.007]}

\newcommand{\tapGapTable}{single layer & 0.284/0.249 & $+0.036$ [+0.007,+0.063] & 0.221/0.244 & $-0.023$ [-0.050,+0.007]\\
$+$ history & 0.232/0.231 & $+0.001$ [-0.025,+0.025] & 0.172/0.227 & $-0.055$ [-0.082,-0.027]\\
3 layers & 0.278/0.296 & $-0.018$ [-0.056,+0.021] & 0.290/0.292 & $-0.002$ [-0.035,+0.032]\\
3 layers $+$ history & 0.299/0.285 & $+0.014$ [-0.023,+0.052] & 0.016/-2.096 & $+2.112$ [+1.832,+2.467]\\}
\newcommand{\tmpDefOW}{5}
\newcommand{\tmpDefN}{240}
\newcommand{\tmpDefRate}{2.1}
\newcommand{\tmpDefCI}{[0.9\%,4.8\%]}
\newcommand{\tmpDefWithin}{+1.1}
\newcommand{\tmpDefWithinCI}{[-4.2,+6.4]}
\newcommand{\tmpDefPeak}{56}
\newcommand{\tmpDefEnd}{37}
\newcommand{\tmpAltOW}{3}
\newcommand{\tmpAltN}{240}
\newcommand{\tmpAltRate}{1.2}
\newcommand{\tmpAltCI}{[0.4\%,3.6\%]}
\newcommand{\tmpAltWithin}{+8.4}
\newcommand{\tmpAltWithinCI}{[+3.3,+14.8]}
\newcommand{\tmpAltPeak}{60}
\newcommand{\tmpAltEnd}{48}
\newcommand{\gDenseOW}{7}

\newcommand{\gMidOW}{5}

\newcommand{\gCoarseOW}{5}

\newcommand{\gDenseN}{240}
\newcommand{\gDenseGained}{2}
\newcommand{\gDenseKept}{5}
\newcommand{\dmDelta}{+0.7}
\newcommand{\dmCI}{[-0.3,+1.8]}
\newcommand{\dmProblems}{120}
\newcommand{\dmOWgreedy}{5}
\newcommand{\dmN}{240}
\newcommand{\dmOWsampled}{6/7/4}
\newcommand{\dmK}{3}
\newcommand{\brCtrlN}{30}

\newcommand{\brMeanPi}{0.80}
\newcommand{\brMeanPhi}{0.16}
\newcommand{\brMeanGap}{+0.64}
\newcommand{\brM}{16}
\newcommand{\brCont}{16}

\newcommand{\fvNoAnsPct}{69}
\newcommand{\brPairsTable}{326 & 2048 & 1.00 & 0.00 & +1.00\\
101 & 1536 & 1.00 & 0.00 & +1.00\\
493 & 1536 & 1.00 & 0.19 & +0.81\\
128 & 512 & 0.81 & 0.00 & +0.81\\
274 & 2048 & 0.94 & 0.12 & +0.81\\
120 & 1536 & 1.00 & 0.25 & +0.75\\
126 & 1024 & 0.62 & 0.00 & +0.62\\
398 & 512 & 0.94 & 0.31 & +0.62\\
381 & 512 & 0.56 & 0.00 & +0.56\\
9 & 2048 & 0.38 & 0.00 & +0.38\\
382 & 2048 & 0.38 & 0.06 & +0.31\\
238 & 1536 & 0.94 & 1.00 & -0.06\\}
\newcommand{\brCapPct}{76}
\newcommand{\brContN}{192}
\newcommand{\brCtrlCapPct}{7}
\newcommand{\brCtrlContN}{480}
\newcommand{\verdictAUCthree}{0.747}
\newcommand{\sevBverdictAUCthree}{0.735}
\newcommand{\decompTotal}{-8.3}
\newcommand{\decompWithin}{+8.5}
\newcommand{\decompComp}{-16.8}
\newcommand{\decompWithinCI}{[+2.1,+14.5]}
\newcommand{\decompCompCI}{[-22.3,-12.0]}
\newcommand{\sevBdecompTotal}{-4.4}
\newcommand{\sevBdecompWithin}{+20.0}
\newcommand{\sevBdecompComp}{-24.4}
\newcommand{\sevBdecompWithinCI}{[+12.6,+27.9]}
\newcommand{\sevBdecompCompCI}{[-31.9,-17.6]}
\newcommand{\decompTable}{512 & 367 & 30.0 & -- & -- & & 262 & 25.6 & -- & --\\
1024 & 358 & 50.6 & +21.5 & -0.9 & & 257 & 44.7 & +20.2 & -1.1\\
1536 & 319 & 56.1 & +11.0 & -5.4 & & 236 & 58.1 & +16.9 & -3.6\\
2048 & 270 & 56.7 & +6.7 & -6.1 & & 197 & 50.8 & -2.5 & -4.8\\
2560 & 222 & 51.8 & +3.6 & -8.5 & & 166 & 51.8 & +7.2 & -6.2\\
3072 & 192 & 50.0 & +2.1 & -3.9 & & 146 & 51.4 & +2.7 & -3.2\\
3584 & 172 & 50.6 & +3.5 & -2.9 & & 125 & 53.6 & +8.0 & -5.8\\
4096 & 151 & 48.3 & -0.7 & -1.6 & & 97 & 46.4 & +2.1 & -9.3\\}
\newcommand{\lenAny}{13}
\newcommand{\lenAnyPct}{3.5}
\newcommand{\lenSus}{8}
\newcommand{\lenSusPct}{2.2}
\newcommand{\lenFlip}{40}
\newcommand{\lenFlipPct}{10.9}
\newcommand{\sevBlenAny}{9}
\newcommand{\sevBlenAnyPct}{3.4}
\newcommand{\sevBlenSus}{3}
\newcommand{\sevBlenSusPct}{1.1}
\newcommand{\sevBlenFlip}{26}
\newcommand{\sevBlenFlipPct}{9.9}
\newcommand{\linRtwo}{0.16}
\newcommand{\mlpRtwo}{0.26}
\newcommand{\sevBlinRtwo}{-1.29}

\newcommand{\thrBestPct}{99}
\newcommand{\thrBestDelta}{+0.0}

\newcommand{\sevBthrBestDelta}{+0.0}
\newcommand{\sweepConfigs}{19}
\newcommand{\sweepSeeds}{10}
\newcommand{\sweepRecMed}{98.6}
\newcommand{\sweepRecMin}{36.9}
\newcommand{\sweepRecMax}{99.5}

\newcommand{\scVoteWithin}{+12.0}
\newcommand{\scVoteComp}{-12.0}
\newcommand{\scVoteWithinCI}{[+4.6,+20.2]}
\newcommand{\scVoteCompCI}{[-17.7,-7.3]}
\newcommand{\scBonWithin}{+5.4}
\newcommand{\scBonComp}{-9.3}
\newcommand{\qwenN}{367}
\newcommand{\qwenOW}{11}
\newcommand{\qwenOWstrict}{5}
\newcommand{\qwenPct}{3.0}
\newcommand{\qwenAcc}{78.2}
\newcommand{\qwenCI}{[1.7\%,5.3\%]}
\newcommand{\qwenTotal}{-2.6}
\newcommand{\qwenWithin}{+7.6}
\newcommand{\qwenComp}{-10.2}
\newcommand{\qwenWithinCI}{[+1.6,+13.7]}
\newcommand{\qwenCompCI}{[-14.3,-6.1]}
\newcommand{\qwenSC}{253}
\newcommand{\qwenLate}{34}
\newcommand{\qwenNever}{69}
\newcommand{\llamaN}{367}
\newcommand{\llamaOW}{8}
\newcommand{\llamaOWstrict}{5}
\newcommand{\llamaPct}{2.2}
\newcommand{\llamaAcc}{78.7}
\newcommand{\llamaCI}{[1.1\%,4.2\%]}
\newcommand{\llamaTotal}{-9.3}
\newcommand{\llamaWithin}{+16.6}
\newcommand{\llamaComp}{-25.9}
\newcommand{\llamaWithinCI}{[+9.3,+24.1]}
\newcommand{\llamaCompCI}{[-31.8,-20.9]}
\newcommand{\llamaSC}{252}
\newcommand{\llamaLate}{37}
\newcommand{\llamaNever}{70}

\newcommand{\phiOW}{8}
\newcommand{\phiOWstrict}{5}
\newcommand{\phiPct}{2.2}
\newcommand{\phiAcc}{82.6}
\newcommand{\phiCI}{[1.1\%,4.2\%]}
\newcommand{\phiTotal}{-7.5}
\newcommand{\phiWithin}{+9.1}
\newcommand{\phiComp}{-16.6}
\newcommand{\phiWithinCI}{[+1.8,+16.6]}
\newcommand{\phiCompCI}{[-22.7,-11.1]}
\newcommand{\phiSC}{272}
\newcommand{\phiLate}{31}
\newcommand{\phiNever}{56}
\newcommand{\blNmethods}{48}
\newcommand{\blNpure}{28}
\newcommand{\blNapprox}{16}
\newcommand{\blNnotimpl}{4}
\newcommand{\blNimpl}{44}
\newcommand{\blNbeatCheaper}{0}
\newcommand{\blAlarmVsFixed}{+0.3}

\newcommand{\blSaveN}{21}
\newcommand{\blSaveMild}{-0.9}
\newcommand{\blSaveWorst}{-45.5}
\newcommand{\blSaveCIzero}{4}
\newcommand{\sevBblNmethods}{28}

\newcommand{\sevBblNimpl}{21}

\newcommand{\sevBblSaveN}{16}
\newcommand{\sevBblSaveMild}{-1.3}
\newcommand{\sevBblSaveWorst}{-47.5}
\newcommand{\sevBblSaveCIzero}{3}
\newcommand{\ntSingleCommit}{0.325}

\newcommand{\ntSingleVerdict}{0.418}

\newcommand{\ntNestedCommit}{0.295}

\newcommand{\ntNestedVerdict}{0.362}

\newcommand{\ntNestedGap}{-0.067}
\newcommand{\sevBntSingleCommit}{0.300}
\newcommand{\sevBntSingleCommitNull}{0.063}
\newcommand{\sevBntSingleVerdict}{0.359}

\newcommand{\sevBntNestedCommit}{0.255}
\newcommand{\sevBntNestedCommitNull}{0.106}
\newcommand{\sevBntNestedVerdict}{0.336}

\newcommand{\sevBntNestedGap}{-0.081}
\newcommand{\nlAggSlope}{-9.1}
\newcommand{\nlWithinSlope}{-8.4}
\newcommand{\nlCompPct}{8}
\newcommand{\nlNatSlope}{-2.4}
\newcommand{\nlNatSlopeCI}{[-4.5,-0.6]}
\newcommand{\nlCensPct}{41}
\newcommand{\nlSurvDelta}{-20.2}
\newcommand{\sevBnlAggSlope}{-8.6}
\newcommand{\sevBnlWithinSlope}{-8.4}
\newcommand{\sevBnlCompPct}{3}
\newcommand{\sevBnlNatSlope}{-1.1}
\newcommand{\sevBnlNatSlopeCI}{[-2.8,+0.3]}
\newcommand{\sevBnlCensPct}{37}
\newcommand{\sevBnlSurvDelta}{-23.7}

\newcommand{\gdSCci}{[+0.0032,+0.0079]}
\newcommand{\consKapDelta}{-0.9\,[-2.7,+0.0]}

\newcommand{\sevBgdSCci}{[+0.0111,+0.0202]}

\newcommand{\probeECE}{0.074}

\newcommand{\verdictECE}{0.138}

\newcommand{\evalOW}{2}
\newcommand{\evalSC}{79}

\newcommand{\gdOWmean}{+0.0144}
\newcommand{\gdSCmean}{+0.0052}
\newcommand{\gdLBmean}{+0.0035}
\newcommand{\gdNCmean}{+0.0087}
\newcommand{\energyNear}{\ensuremath{1.5\times10^{-3}}}
\newcommand{\energyFar}{\ensuremath{1.6\times10^{-3}}}
\newcommand{\ncLoneTwoRate}{1.5}
\newcommand{\ncLthreeFiveRate}{2.7}

\newcommand{\ncMatchedLoneTwo}{\ensuremath{1.5\times10^{-3}}}
\newcommand{\ncMatchedLthreeFive}{\ensuremath{1.7\times10^{-3}}}

\newcommand{\budgetDeltaCapped}{-0.7\,[-4.0,+2.6]}
\newcommand{\cappedN}{151}
\newcommand{\voteAcc}{71.8}

\newcommand{\voteTokens}{3087}
\newcommand{\agreeAcc}{65.5}
\newcommand{\agreeTokens}{1625}
\newcommand{\fullAcc}{77.3}
\newcommand{\fullTokens}{2808}
\newcommand{\alarmAcc}{74.5}

\newcommand{\alarmTokens}{2624}
\newcommand{\alarmAccOW}{0.0}
\newcommand{\alarmFireSC}{13}
\newcommand{\alarmFireOW}{0}
\newcommand{\alarmAccSC}{97.5}
\newcommand{\opAccLB}{83.3}
\newcommand{\opAccNC}{0.0}
\newcommand{\opAccOWstrict}{0.0}

\newcommand{\sevBpopOverthink}{9}

\newcommand{\sevBevalOW}{2}
\newcommand{\sevBevalSC}{50}

\newcommand{\sevBgdOWmean}{+0.0228}
\newcommand{\sevBgdSCmean}{+0.0154}
\newcommand{\sevBgdLBmean}{+0.0203}
\newcommand{\sevBgdNCmean}{+0.0240}
\newcommand{\sevBenergyNear}{\ensuremath{7.0\times10^{-3}}}
\newcommand{\sevBenergyFar}{\ensuremath{7.1\times10^{-3}}}

\newcommand{\sevBbudgetDeltaCapped}{+2.1\,[-1.0,+6.2]}
\newcommand{\sevBcappedN}{97}
\newcommand{\sevBvoteAcc}{65.0}

\newcommand{\sevBvoteTokens}{3315}
\newcommand{\sevBagreeAcc}{61.3}
\newcommand{\sevBagreeTokens}{1535}
\newcommand{\sevBfullAcc}{77.5}
\newcommand{\sevBfullTokens}{3015}
\newcommand{\sevBalarmAcc}{73.8}

\newcommand{\sevBalarmTokens}{2767}

\newcommand{\sevBalarmFireSC}{14}
\newcommand{\sevBalarmFireOW}{50}

}{}
\newcommand{\matchedCaseAcc}{50.0}
\newcommand{\matchedControlAcc}{97.9}
\newcommand{\matchedCaseGap}{50.0}
\newcommand{\matchedControlGap}{2.1}
\newcommand{\matchedGapContrast}{47.9}
\newcommand{\matchedGapContrastCI}{[27.1, 69.8]}
\newcommand{\matchedCaseGapCI}{[28.1, 71.9]}
\newcommand{\matchedControlGapCI}{[0.0, 6.2]}

\newcommand{\matchedCaseCorrect}{48}
\newcommand{\matchedControlCorrect}{94}
\newcommand{\refitCasePositive}{4}
\newcommand{\refitControlPositive}{1}

\newcommand{\refitSmallGapMean}{-0.088}

\newcommand{\refitLargeGapMean}{-0.029}
\newcommand{\matchedPairRows}{%
9/109 & 5 & 2048 & 1/8 & $[0.02, 0.47]$ & 8/8 & $[0.68, 1.00]$ \\
101/63 & 5 & 1536 & 6/8 & $[0.41, 0.93]$ & 8/8 & $[0.68, 1.00]$ \\
120/57 & 2 & 1536 & 7/8 & $[0.53, 0.98]$ & 8/8 & $[0.68, 1.00]$ \\
126/461 & 4 & 1024 & 2/8 & $[0.07, 0.59]$ & 6/8 & $[0.41, 0.93]$ \\
128/471 & 3 & 512 & 0/8 & $[0.00, 0.32]$ & 8/8 & $[0.68, 1.00]$ \\
238/459 & 5 & 1536 & 7/8 & $[0.53, 0.98]$ & 8/8 & $[0.68, 1.00]$ \\
274/328 & 5 & 2048 & 8/8 & $[0.68, 1.00]$ & 8/8 & $[0.68, 1.00]$ \\
326/35 & 2 & 2048 & 1/8 & $[0.02, 0.47]$ & 8/8 & $[0.68, 1.00]$ \\
381/313 & 5 & 512 & 0/8 & $[0.00, 0.32]$ & 8/8 & $[0.68, 1.00]$ \\
382/232 & 5 & 2048 & 2/8 & $[0.07, 0.59]$ & 8/8 & $[0.68, 1.00]$ \\
398/73 & 4 & 512 & 6/8 & $[0.41, 0.93]$ & 8/8 & $[0.68, 1.00]$ \\
493/323 & 5 & 1536 & 8/8 & $[0.68, 1.00]$ & 8/8 & $[0.68, 1.00]$ \\
}
\newcommand{\refitDetailRows}{%
1 & 0.2201 & 0.2332 & -0.0130 & 0.2254 & 0.2342 & -0.0088 \\
2 & -0.6384 & 0.2512 & -0.8896 & 0.1837 & 0.2427 & -0.0590 \\
3 & 0.2372 & 0.2553 & -0.0180 & 0.2403 & 0.2161 & 0.0242 \\
4 & 0.3045 & 0.2790 & 0.0255 & 0.2193 & 0.2546 & -0.0353 \\
5 & 0.2449 & 0.2539 & -0.0090 & 0.2161 & 0.2437 & -0.0276 \\
6 & 0.2194 & 0.1964 & 0.0230 & 0.1857 & 0.1862 & -0.0005 \\
7 & 0.2702 & 0.2460 & 0.0242 & 0.1804 & 0.2774 & -0.0970 \\
8 & 0.1921 & 0.2532 & -0.0611 & 0.1857 & 0.2227 & -0.0370 \\
9 & 0.2999 & 0.1953 & 0.1047 & 0.2368 & 0.2458 & -0.0090 \\
10 & 0.1708 & 0.2347 & -0.0639 & 0.2095 & 0.2484 & -0.0389 \\
}

\newcommand{\cenN}{379}
\newcommand{\cenNlvl}{255}
\newcommand{\cenCont}{96.4}
\newcommand{\cenLoss}{3.6}
\newcommand{\cenLossCI}{[2.2, 5.2]}
\newcommand{\cenLossLvl}{4.2}
\newcommand{\cenLossLvlCI}{[2.5, 6.2]}
\newcommand{\cenHalfPct}{3.4}
\newcommand{\cenHalf}{13}
\newcommand{\cenZero}{3}

\newcommand{\cenCaseCont}{47.9}
\newcommand{\cenOtherCont}{98.0}
\newcommand{\cenKeepAll}{348}
\newcommand{\cenAnyLoss}{31}
\newcommand{\injN}{96}
\newcommand{\injImm}{91.7}
\newcommand{\injCont}{95.8}
\newcommand{\injCtrl}{94.8}
\newcommand{\injDrop}{+1.0}
\newcommand{\injDropCI}{[-1.0, +3.1]}
\newcommand{\hintN}{90}
\newcommand{\hintCont}{93.6}
\newcommand{\hintCtrl}{94.4}
\newcommand{\hintDrop}{0.8}
\newcommand{\hintDropCI}{[-1.7, 3.6]}
\newcommand{\hintAdopt}{5.0}
\newcommand{\hintImm}{32.2}
\newcommand{\hintImmAdopt}{64.4}

\newcommand{\hintModel}{60}
\newcommand{\hintPlus}{30}
\newcommand{\hintExcl}{6}
\newcommand{\propOD}{2.6}
\newcommand{\propICC}{0.23}
\newcommand{\propHigh}{5}
\newcommand{\propTopTen}{40}
\newcommand{\propAny}{57}
\newcommand{\propAnyNull}{89}
\newcommand{\dfmOW}{9}
\newcommand{\dfmN}{262}
\newcommand{\dfmPct}{3.4}
\newcommand{\dfmCI}{[1.8, 6.4]}
\newcommand{\dfmWithin}{+7.3}
\newcommand{\dfmWithinCI}{[+0.1, +14.2]}
\newcommand{\dfmComp}{-13.5}
\newcommand{\dfmTotal}{-6.3}
\newcommand{\orcDelta}{+1.8}
\newcommand{\orcDeltaCI}{[+0.0, +4.5]}
\newcommand{\orcSaved}{39}
\newcommand{\orcAcc}{79.1}
\newcommand{\orcTokens}{1713}
\newcommand{\sevBorcDelta}{+2.5}
\newcommand{\sevBorcDeltaCI}{[+0.0, +6.2]}

\newcommand{\sevBorcAcc}{80.0}
\newcommand{\sevBorcTokens}{1836}

\newcommand{\lbShortPct}{2.5}
\newcommand{\lbShortCI}{[1.3, 4.6]}
\newcommand{\lbShortCapped}{44}
\newcommand{\lbShortAcc}{75.7}
\newcommand{\lbLongN}{367}
\newcommand{\lbLongPct}{3.3}
\newcommand{\lbLongCI}{[1.9, 5.6]}
\newcommand{\lbLongCapped}{11}
\newcommand{\lbLongAcc}{80.9}
\newcommand{\lbWithin}{+7.9}
\newcommand{\lbWithinCI}{[-4.4, +21.8]}
\newcommand{\lbComp}{-36.5}
\newcommand{\lbTotal}{-28.6}
\newcommand{\nfOW}{8}
\newcommand{\nfN}{262}
\newcommand{\nfPct}{3.1}
\newcommand{\nfCI}{[1.6, 5.9]}
\newcommand{\nfWithin}{+15.0}
\newcommand{\nfComp}{-17.2}
\newcommand{\nfTotal}{-2.2}
\newcommand{\nfWithinCI}{[+8.4, +22.1]}
\newcommand{\nfAcc}{71.4}
\newcommand{\dfmAcc}{69.8}
\newcommand{\obOW}{14}
\newcommand{\obN}{367}
\newcommand{\obPct}{3.8}
\newcommand{\obCI}{[2.3, 6.3]}
\newcommand{\obAcc}{48.8}
\newcommand{\obTotal}{-4.4}
\newcommand{\obWithin}{+7.3}
\newcommand{\obWithinCI}{[+3.1, +11.8]}
\newcommand{\obComp}{-11.7}
\newcommand{\ieOW}{8}
\newcommand{\ieN}{262}
\newcommand{\iePct}{3.1}
\newcommand{\ieCI}{[1.6, 5.9]}
\newcommand{\ieAcc}{78.6}
\newcommand{\ieTotal}{-4.1}
\newcommand{\ieWithin}{+17.9}
\newcommand{\ieWithinCI}{[+10.0, +26.0]}
\newcommand{\ieComp}{-22.0}
\newcommand{\qfOW}{7}
\newcommand{\qfN}{367}
\newcommand{\qfPct}{1.9}
\newcommand{\qfCI}{[0.9, 3.9]}
\newcommand{\qfAcc}{66.2}
\newcommand{\qfTotal}{+7.1}
\newcommand{\qfWithin}{+8.7}
\newcommand{\qfWithinCI}{[+5.6, +11.9]}
\newcommand{\qfComp}{-1.7}

\providecommand{\placeholderfig}[2]{\fbox{\parbox[c][#2][c]{0.94\linewidth}{\centering
  \texttt{figure pending: \detokenize{#1}}}}}
\newcommand{\maybefig}[3]{\IfFileExists{#1}{\includegraphics[width=#2]{#1}}{\placeholderfig{#1}{#3}}}

\title{Conserved Verdicts, Drifting Commitments:\\
Identifying Harmful Overthinking}
\author{Jianghong Zhou \And Hao Rong \And Chen Lin}

\begin{document}
\maketitle
\lhead{Preprint}

\begin{abstract}
A reasoning model that has reached a correct answer can lose it by continuing
to think. Deciding whether to stop means comparing two probabilities at the
same point of a chain: that the answer the model would give if stopped now is
correct (its \emph{commitment}), and that the answer it will finish with is
correct (its \emph{verdict}). We show that the verdict alone, which is what
final-correctness probes estimate, cannot make this comparison. As a forecast
of a fixed outcome it has no predictable drift, whether continuing helps or
hurts. The commitment--verdict gap equals the expected remaining change in
answer quality and the accuracy gained by stopping. A second identity splits
any accuracy-versus-budget curve into answer change within chains and change in
which chains are still running. We measure both on MATH-500. At all $\cenN$
points where DeepSeek-R1-Distill-Qwen-1.5B holds a correct answer, selected
without regard to how its chain ends, continuing loses $\cenLoss$ points on
average (95\% CI $\cenLossCI$), and the loss concentrates in a few problems.
In six reasoning models from three families, on MATH-500 and OlympiadBench and
with budgets up to 16{,}384 tokens, declining accuracy curves come from which
chains remain, while the chains still reasoning keep or improve their answers. An oracle
stopping rule gains at most $\orcDelta$ and $\sevBorcDelta$ points at 1.5B and
7B. The framework says what to measure before building a stopping rule.
\end{abstract}

\section{Introduction}
\label{sec:intro}

Reasoning models solve hard problems by thinking longer, and spending more
inference-time computation is now a main way to improve them
\citep{wei2022chain,snell2024scaling,muennighoff2025s1}. Longer is not always
better. A model can reach a correct answer and then talk itself out of it
\citep{wei2025evolution}, and accuracy is often reported to fall at long
budgets \citep{zhou2026morethinking}. Early-exit methods try to save tokens and
keep such answers \citep{zhang2025reasoning,liu2025answerconvergence,fu2025certaindex}.
All of them face one question at every point of a chain:
\emph{is it better to stop now, or to keep reasoning?}

The question compares two numbers at the same point. The \textbf{commitment}
is the probability that the answer the model would give if stopped now is
correct. The \textbf{verdict} is the probability that the answer it will end
with, if it keeps reasoning, is correct. Stopping helps exactly when the
commitment is the larger of the two.

\begin{figure}[t]
\centering
\includegraphics[width=\textwidth]{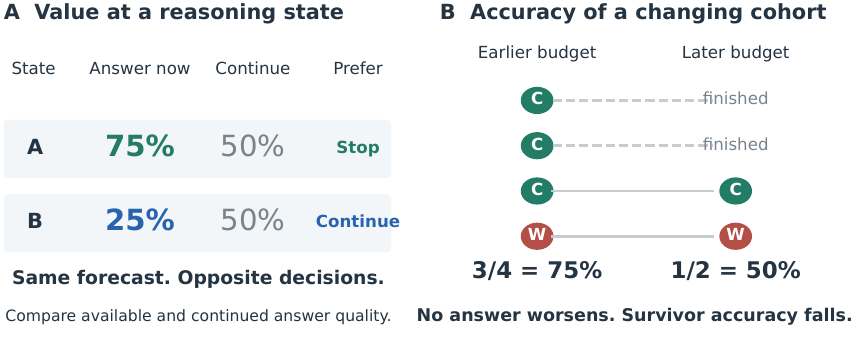}
\caption{\textbf{Two comparisons determine what further reasoning buys.}
Illustrative numbers (C: correct; W: wrong).
\textbf{Left:} two chains have the same 50\% chance of ending correct, yet the
answer available now is right 75\% of the time in one and 25\% in the other,
so stopping helps one and hurts the other.
\textbf{Right:} two correct chains finish early; accuracy among the chains
still running falls from 75\% to 50\% although no answer changes.
Comparing the same chains over time separates answer change from selection.}
\label{fig:idmap}
\end{figure}

Two common measurements answer different questions
(Fig.~\ref{fig:idmap}). First, probes and value models trained to predict
whether a chain will end correct estimate the verdict. However accurate, the
verdict cannot decide when to stop on its own. Two chains can both have a 50\%
chance of ending correct while their current answers are right 75\% and 25\%
of the time. Stopping helps the first chain and hurts the second.
Second, accuracy-versus-budget curves usually average over the chains still
running at each budget. Easy problems tend to finish early and correctly, so
the chains left at long budgets are harder ones. Accuracy among them can fall
even when no chain's answer gets worse.

\paragraph{This paper.} We make both comparisons exact and then measure them.
\begin{itemize}[leftmargin=1.2em,itemsep=1pt,topsep=2pt]
\item \emph{The verdict is conserved.} For a fixed way of continuing, an exact
forecast of the final outcome cannot predict its own next revision; it is a
martingale. A stable final-correctness forecast is therefore no evidence that
reasoning is harmless (Prop.~\ref{prop:verdict}, Cor.~\ref{cor:nonidentification}).
\item \emph{The gap is the value of stopping.} The commitment--verdict gap
equals the expected remaining change in the commitment, and equals the accuracy
gained by stopping (Eq.~\ref{eq:gap}, Cor.~\ref{cor:advantage}).
\item \emph{Budget curves mix two effects.} An exact identity splits any change
in a survivor-averaged curve into answer change within chains and change in
which chains remain (Prop.~\ref{prop:decomp}).
\end{itemize}
These results use standard probability: the tower property, the Doob
decomposition and telescoping. The commitment--verdict gap is the familiar
comparison of stopping and continuing values in optimal stopping
\citep{chow1971great}. What we add is which quantities answer the overthinking
question, why two common practices measure something else, and the
measurements that answer it. On MATH-500 \citep{hendrycks2021math} and
OlympiadBench \citep{he2024olympiadbench} we find:
\begin{itemize}[leftmargin=1.2em,itemsep=1pt,topsep=2pt]
\item \emph{Losing a correct answer is real, rare, and concentrated.} At all
$\cenN$ points where the 1.5B model holds a correct answer, chosen without
looking at how its chain ends, fresh continuations lose $\cenLoss$ points on
average. In $\cenKeepAll$ of the $\cenN$ points the model never loses the
answer; a few problems account for most losses. A misleading hint inserted
mid-reasoning makes the stop-now answer wrong in most cases, yet continuing
restores it: the commitment collapses while the verdict holds (Sec.~\ref{sec:exp1}).
\item \emph{Declining budget curves come from selection.} In six models from
three families, on two benchmarks, at several numeric precisions and with budgets up
to 16{,}384 tokens, chains still reasoning keep or improve their answers; the
decline comes from which chains remain (Sec.~\ref{sec:exp1b}).
\item \emph{Early stopping has little room to gain.} An oracle that stops each
chain at its best checkpoint gains $\orcDelta$ points at 1.5B and
$\sevBorcDelta$ at 7B. None of the replayed early-exit rules improves accuracy
at lower token cost (Sec.~\ref{sec:stopping}).
\end{itemize}

\section{Commitment, Verdict, and Drift}
\label{sec:setup}

Consider one reasoning chain on one problem, continued by a fixed protocol: the
model keeps sampling until it answers or reaches its token budget. Let
$\F_t$ denote everything generated up to step $t$, and $Y\in\{0,1\}$ whether
the final answer is correct. A step can be a paragraph of reasoning or a fixed
number of tokens.

\paragraph{Two readouts.} The \emph{commitment} $\pi_t$ is the probability
that the answer elicited at step $t$, by a fixed stop-and-answer procedure, is
correct. The \emph{verdict} is $\Phi_t=\E[Y\mid\F_t]$, the probability that the
final answer will be correct. When the chain ends at step $\tau$, the current
answer is the final answer, so $\pi_\tau=Y=\Phi_\tau$. The formal assumptions
(a Markov reasoning state and a terminal answer read from the final state) are
in App.~\ref{app:proofs}.

\paragraph{Drift.} Split each change in the commitment into an expected part
and a surprise:
\begin{equation}
  \Delta_t := \pi_{t+1}-\pi_t = b_t + \eta_t,\qquad
  b_t := \E[\Delta_t\mid\F_t],\quad \E[\eta_t\mid\F_t]=0.
  \label{eq:split}
\end{equation}
The \emph{drift} $b_t$ is positive when continuing is expected to improve the
current answer and negative when it is expected to erode it. The surprise
$\eta_t$ averages to zero. This split is unique (Thm.~\ref{thm:doob}), so
drift is a well-defined property of the chain and the elicitation procedure.

\section{What Identifies the Value of Continuing}
\label{sec:doob}

\begin{proposition}[Conserved verdict]
\label{prop:verdict}
\begin{enumerate}[leftmargin=1.5em,itemsep=1pt,topsep=1pt]
\item[(i)] The verdict is a martingale: $\E[\Phi_{t+1}\mid\F_t]=\Phi_t$ for
every $t$, under any fixed continuation protocol.
\item[(ii)] If the chain ends at $\tau$ with $\E[\tau]<\infty$ (automatic
under a token budget) and $\pi_\tau=Y$, then
\begin{equation}
  \pi_t-\Phi_t \;=\; -\,\E\Big[\textstyle\sum_{s\ge t}b_s\,\Big|\,\F_t\Big].
  \label{eq:gap}
\end{equation}
In particular $\pi_t=\Phi_t$ at every step if and only if the drift is zero
throughout.
\end{enumerate}
\end{proposition}

\noindent\emph{Proof sketch.} Part (i) is the tower property. For (ii), add up
the expected changes $\E[b_s\mid\F_t]$ from $t$ to the end and use
$\pi_\tau=Y$. App.~\ref{app:proofs} gives the full proof.

Part (i) is why a final-correctness forecast cannot reveal harm. An accurate
forecast already averages over what continuing will do, so its next revision
is unpredictable whether continuing helps or hurts. Part (ii) says the gap
between the two readouts is exactly the drift still to come: if continuing will
on balance erode the current answer, the commitment sits above the verdict by
that amount.

\begin{corollary}[A verdict trajectory does not decide whether to stop]
\label{cor:nonidentification}
Two chains can share the same verdict at every step and still have opposite
signs of $\pi_t-\Phi_t$.
\end{corollary}
\noindent\emph{Proof.} Let the chain end at $\tau=2$ and reveal
$Y\sim\mathrm{Bernoulli}(1/2)$ only then. Set $\pi_0=1/2$, $\pi_1=1/2\pm\epsilon$
and $\pi_2=Y$. Both versions have $\Phi_0=\Phi_1=1/2$, but stopping at step 1
gains $\pm\epsilon$. With $\epsilon=1/4$ this is the left panel of
Fig.~\ref{fig:idmap}. \hfill$\square$

\begin{corollary}[Stopping advantage]
\label{cor:advantage}
Let $S\le\tau$ be a stopping rule that uses only $\F_S$, and $C_S\in\{0,1\}$
the correctness of the answer it returns, with $\E[C_S\mid\F_S]=\pi_S$. Then
\begin{equation}
 \E[C_S-Y]=\E[\pi_S-\Phi_S]
 =-\E\!\left[\sum_{t=S}^{\tau-1}b_t\right].
 \label{eq:advantage}
\end{equation}
\end{corollary}
\noindent\emph{Proof.} Apply the tower property to $C_S$ and $Y$, then
Eq.~\eqref{eq:gap} at $S$. \hfill$\square$

A stopping rule gains accuracy only from the states where it stops and where
remaining drift is negative. Its best possible gain is the total expected loss
at such states, which we measure directly in Sec.~\ref{sec:exp1}. If continuing
also has a cost $K_S$ priced at $\lambda$ per unit, the value of continuing is
$\Phi_S-\pi_S-\lambda\,\E[K_S\mid\F_S]$, so an efficiency-driven exit can be
right even when continuing would improve accuracy. App.~\ref{app:operator}
writes the gap as a Poisson equation for Markov reasoning states, and
App.~\ref{app:objective} turns the drift into training objectives.

\paragraph{From one chain to a population.}
Accuracy-versus-budget curves average a correctness indicator $A(T)$, the
correctness of the forced answer at budget $T$, over the chains still running.
Write $M(T)=\E[A(T)\mid\tau\ge T]$.

\begin{proposition}[Survivorship decomposition]\label{prop:decomp}
For budgets $T<T'$,
\begin{equation*}
M(T')-M(T)=\underbrace{\E\!\left[A(T')-A(T)\mid\tau\ge T'\right]}_{\textnormal{within-cohort change}}
+\underbrace{\E\!\left[A(T)\mid\tau\ge T'\right]-\E\!\left[A(T)\mid\tau\ge T\right]}_{\textnormal{composition change}}.
\end{equation*}
\end{proposition}
\noindent\emph{Proof.} Add and subtract $\E[A(T)\mid\tau\ge T']$. \hfill$\square$

The within-cohort term compares the same chains at two budgets: it is the rate
of wrong-to-correct minus correct-to-wrong changes. The composition term is the
effect of chains finishing. Only the first measures what longer reasoning does
to answers.

\paragraph{Two measurements.} The results translate into two experiments. At a
fixed prefix, compare the answer elicited now with fresh continuations from
the same prefix; their difference estimates $\pi_t-\Phi_t$. Across a
population, compute the decomposition on fixed cohorts.

\section{Experiments}
\label{sec:experiments}

\subsection{Setup}
\label{sec:realsetup}

The primary configurations are DeepSeek-R1-Distill-Qwen-1.5B in FP16 on 367
MATH-500 problems (levels 3--5) and DeepSeek-R1-Distill-Qwen-7B on 262 problems
(levels 4--5) \citep{deepseek2025r1,hendrycks2021math}. The 7B model is loaded
in 4-bit NF4 \citep{dettmers2023qlora}; an 8-bit run \citep{dettmers2022llmint8}
gives the same results (Sec.~\ref{sec:exp1b}). Each
problem gets one chain sampled at temperature $0.6$ and top-$p$ $0.95$, up to
4096 tokens. Every 512 tokens, a separate branch closes the thinking block and
asks for a boxed answer, decoded greedily for at most 48 tokens
\citep{muennighoff2025s1}; the chain itself continues unchanged. Answers are
scored by normalized string matching and symbolic verification.
A chain is \emph{early-correct/final-wrong} if an answer at or before 2048
tokens is correct and its final answer is wrong. Intervals are 95\% bootstrap
intervals over whole chains or problems unless noted. App.~\ref{app:expdetails}
gives full details, and Table~\ref{tab:status} in App.~\ref{app:rebuttal}
marks which analyses were fixed in advance in a dated plan, which are
follow-ups, and which were added during review.

\subsection{How often does continuing lose a correct answer?}
\label{sec:exp1}

\paragraph{Outcome-blind continuations.} We take every 1.5B chain that has a
correct answer at a checkpoint at or before 2048 tokens, at its first such
checkpoint. This selects $\cenN$ points across all five MATH levels, and the
rule never looks at how a chain ends. From each point we sample four fresh
continuations with the original prefix, temperature and remaining budget, and
score them as in the primary runs. The immediate answer is correct at every
point, so the drop in accuracy from continuing estimates $\pi_t-\Phi_t$.

\begin{figure}[t]
\centering
\includegraphics[width=\textwidth]{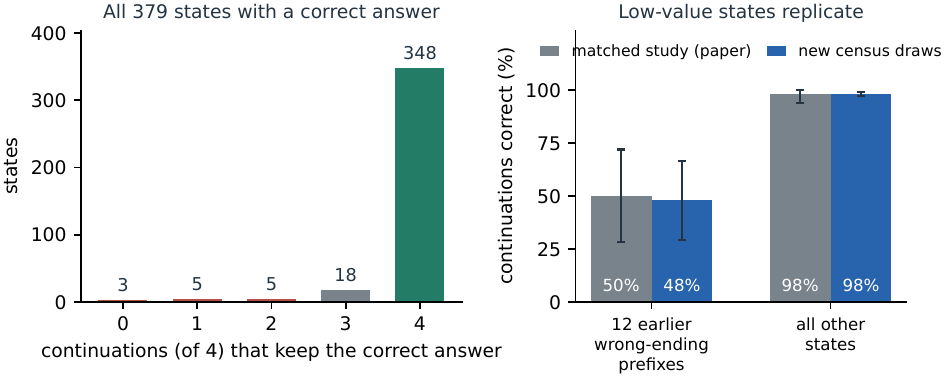}
\caption{\textbf{Continuing rarely loses a correct answer, and the losses
concentrate.} \textbf{Left:} at each of the $\cenN$ points where the 1.5B
model holds a correct answer, how many of four fresh continuations keep it.
\textbf{Right:} continuation accuracy from the 12 prefixes whose original
chains ended wrong, and from all other points. Gray: the eight-draw matched
study; blue: four new draws per point. Bars show 95\% intervals.}
\label{fig:census}
\end{figure}

Continuing loses $\cenLoss$ points on average (95\% CI $\cenLossCI$; levels
3--5: $\cenLossLvl$, $\cenLossLvlCI$). The loss is concentrated
(Fig.~\ref{fig:census}, left): in $\cenKeepAll$ of $\cenN$ points all four
continuations keep the answer, while $\cenHalf$ points ($\cenHalfPct\%$) lose
it in at least two of four. This is in line with the realized rate of
early-correct/final-wrong chains: eight fresh chains for each of 367 problems give $\rEightOW/\rEightN$
such chains ($\rEightRate\%$, problem-clustered CI $\rEightCI$).

\paragraph{Losses sit in particular problems.} The 12 points whose original
chains ended wrong keep their answer in only $\cenCaseCont\%$ of new
continuations, against $\cenOtherCont\%$ for all other points; an earlier
eight-draw matched study gave $\matchedCaseAcc\%$ against
$\matchedControlAcc\%$ (Fig.~\ref{fig:census}, right; App.~\ref{app:matched}).
The repeated-sample data show the same concentration at the level of
problems. If every problem had the same transition rate, the number of
early-correct/final-wrong chains per problem would be binomial. The variance
is $\propOD$ times larger, and $\propHigh$ problems show the transition in at
least five of eight fresh chains, where the binomial predicts none. The ten
most affected problems produce $\propTopTen\%$ of all such chains.

\paragraph{Measurement choices.} The count of early-correct/final-wrong chains
depends on how often answers are elicited. On a matched set of $\gDenseN$
chains it is $\gCoarseOW$, $\gMidOW$ and $\gDenseOW$ at 1024-, 512- and
256-token spacing, so we report rates at a fixed spacing. The forcing string
matters little: where a truncated chain also answers without it, the two
answers agree on correctness in $\fvCorrAgree\%$ of cases, and greedy versus
sampled forcing changes accuracy by $\dmDelta$ points ($\dmCI$). Across
final-answer definitions and at temperature 0.9 the rate stays between 1\% and
4\% (App.~\ref{app:expdetails}).

\paragraph{A misleading hint separates the two readouts.} To push the model
toward abandoning a correct answer, we insert a wrong answer into the reasoning
at $\hintN$ of these points: ``Wait, I made an error earlier. Rechecking
carefully, the correct answer should be~$W$,'' where $W$ is a wrong answer the
model itself gave on that problem, or the integer next to the gold answer. The
hint moves the commitment sharply: an answer forced right after it is correct
in only $\hintImm\%$ of cases and equals $W$ in $\hintImmAdopt\%$. The verdict
barely moves: continuations end correct in $\hintCont\%$ of cases, against
$\hintCtrl\%$ at the same points without the hint (paired drop $\hintDrop$
points, $\hintDropCI$), and end at $W$ in only $\hintAdopt\%$. Here the right
decision is to keep reasoning. A rule that reads only the current answer would
stop at the worst moment, and a final-correctness forecast would not register
the abandonment at all (App.~\ref{app:rebuttal}).

\subsection{Does the declining budget curve reflect worse answers?}
\label{sec:exp1b}

\begin{figure}[t]
\centering
\begin{minipage}[t]{0.46\textwidth}
\centering
\maybefig{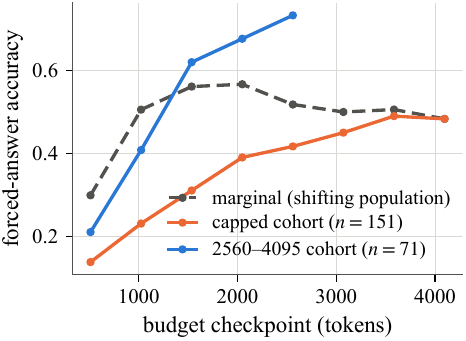}{\linewidth}{2.5cm}
\end{minipage}\hfill
\begin{minipage}[t]{0.46\textwidth}
\centering
\maybefig{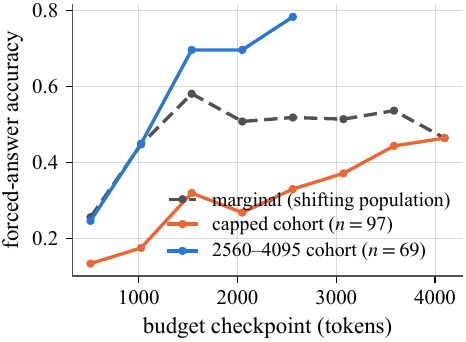}{\linewidth}{2.5cm}
\end{minipage}
\caption{\textbf{Answers improve beneath a declining accuracy curve.}
R1-Distill-Qwen 1.5B, levels 3--5 (\textbf{left}), and 7B, levels 4--5
(\textbf{right}). Gray: accuracy among chains still running at each budget.
Colored: fixed groups of chains followed across budgets; their final-step
changes are $\budgetDeltaCapped$ ($n=\cappedN$) and $\sevBbudgetDeltaCapped$
($n=\sevBcappedN$) points.}
\label{fig:budget}
\end{figure}

Both primary configurations show the familiar decline: accuracy among
unfinished chains falls after about 2000 tokens (Fig.~\ref{fig:budget}, gray).
Applying Prop.~\ref{prop:decomp} over the four 512-token intervals from 2048
to 4096 tokens separates its sources. The chains still reasoning improve, by
$\decompWithin$ points at 1.5B ($\decompWithinCI$) and $\sevBdecompWithin$ at
7B ($\sevBdecompWithinCI$). The decline comes from composition:
$\decompComp$ and $\sevBdecompComp$ points, as chains that finish early are
disproportionately correct. Fixed cohorts followed across all budgets are flat
or rising (Fig.~\ref{fig:budget}, colored).

\begin{table}[t]
\centering\footnotesize
\caption{\textbf{The same pattern across models, benchmarks, difficulty,
precision and budget.} MATH-500 unless marked. Accuracy changes in percentage
points, summed over 512-token intervals
from 2048 to 4096 tokens ($^\dagger$from 4096 to 16{,}384 tokens); brackets
are 95\% chain-bootstrap intervals.}
\label{tab:audit}
\setlength{\tabcolsep}{4pt}
\begin{tabular}{@{}llrrrl@{}}
\toprule
Model & Setting & $n$ & Survivor change & Within-cohort & Composition\\
\midrule
R1-Qwen 1.5B & FP16, levels 3--5 & 367 & $\decompTotal$ & $\decompWithin$ $\decompWithinCI$ & $\decompComp$\\
R1-Qwen 1.5B & FP16, levels 4--5 & $\dfmN$ & $\dfmTotal$ & $\dfmWithin$ $\dfmWithinCI$ & $\dfmComp$\\
R1-Qwen 1.5B & NF4, levels 4--5 & $\nfN$ & $\nfTotal$ & $\nfWithin$ $\nfWithinCI$ & $\nfComp$\\
R1-Qwen 7B & NF4, levels 4--5 & 262 & $\sevBdecompTotal$ & $\sevBdecompWithin$ $\sevBdecompWithinCI$ & $\sevBdecompComp$\\
R1-Qwen 7B & 8-bit, levels 4--5 & $\ieN$ & $\ieTotal$ & $\ieWithin$ $\ieWithinCI$ & $\ieComp$\\
Qwen3 1.7B & FP16, levels 3--5 & 367 & $\qwenTotal$ & $\qwenWithin$ $\qwenWithinCI$ & $\qwenComp$\\
R1-Llama 8B & NF4, levels 3--5 & 367 & $\llamaTotal$ & $\llamaWithin$ $\llamaWithinCI$ & $\llamaComp$\\
Phi-4-mini 3.8B & FP16, levels 3--5 & 367 & $\phiTotal$ & $\phiWithin$ $\phiWithinCI$ & $\phiComp$\\
Qwen3.5 2B & FP16, levels 3--5 & $\qfN$ & $\qfTotal$ & $\qfWithin$ $\qfWithinCI$ & $\qfComp$\\
R1-Qwen 1.5B & FP16, OlympiadBench & $\obN$ & $\obTotal$ & $\obWithin$ $\obWithinCI$ & $\obComp$\\
R1-Qwen 1.5B & FP16, 16k cap$^\dagger$ & $\lbLongN$ & $\lbTotal$ & $\lbWithin$ $\lbWithinCI$ & $\lbComp$\\
\bottomrule
\end{tabular}
\end{table}

\paragraph{Across models and settings.} Table~\ref{tab:audit} repeats the
decomposition for four more reasoning models (Qwen3-1.7B,
R1-Distill-Llama-3.1-8B, Phi-4-mini-reasoning and the 2026 Qwen3.5-2B
\citep{qwen2026qwen35} with thinking enabled), on OlympiadBench, at other
precisions, and with a longer budget. In every configuration the within-cohort
term is positive and the composition term negative; wherever survivor accuracy
falls, composition is the cause. Holding difficulty fixed separates the two
primary configurations: at levels 4--5 the 1.5B model shows the same
early-correct/final-wrong rate as the 7B model ($\dfmOW/\dfmN=\dfmPct\%$
against $\sevBpopOverthink/262$). Precision does not move the rate: $\nfPct\%$
for the 1.5B model in NF4, and $\iePct\%$ ($\ieCI$) for the 7B model in 8-bit.
The rate is $\obPct\%$ ($\obCI$) on OlympiadBench, where final accuracy is only
$\obAcc\%$, and $\qfPct\%$ ($\qfCI$) for Qwen3.5-2B. With a 16{,}384-token
budget, the share of chains stopped by the cap falls from $\lbShortCapped\%$ to
$\lbLongCapped\%$, and the rate is $\lbLongPct\%$ ($\lbLongCI$) against
$\lbShortPct\%$ on the same chains truncated at 4096 tokens. Majority-vote
curves show the same structure (within-cohort $\scVoteWithin$ points,
composition $\scVoteComp$; App.~\ref{app:audit2}).

\paragraph{Accuracy by natural length.} A forcing-free check gives the same
picture. Sorting chains by how long they ran, accuracy falls by
$\nlAggSlope$ (1.5B) and $\sevBnlAggSlope$ (7B) points per 512 tokens. Among
chains that ended on their own, the slopes shrink to $\nlNatSlope$
($\nlNatSlopeCI$) and $\sevBnlNatSlope$ ($\sevBnlNatSlopeCI$); the longest bins
are filled with chains stopped by the token cap ($\nlCensPct\%$ and
$\sevBnlCensPct\%$ of all chains).

\subsection{How much can early stopping gain?}
\label{sec:stopping}

Corollary~\ref{cor:advantage} bounds what any stopping rule can gain by the
expected loss at the states where it stops. We replay stopping rules on
held-out chains, using only information available at each checkpoint, and
charge each forced-answer branch 48 tokens (Table~\ref{tab:stopmain}).
An oracle that knows the correct answer and stops each chain at its best
checkpoint gains $\orcDelta$ points at 1.5B ($\orcDeltaCI$) and
$\sevBorcDelta$ at 7B ($\sevBorcDeltaCI$) while saving $\orcSaved\%$ of tokens.
No real rule can do better, so in this regime early stopping is worth about two
points of accuracy at most, and its practical value is the saved tokens.

\begin{table}[t]
\centering\footnotesize
\caption{\textbf{Stopping replays on held-out chains:} accuracy (\%) and mean
charged tokens. There are 110/80 evaluation chains at 1.5B/7B.
App.~\ref{app:diagnostics} reports all $\blNmethods$ configurations from the
published method families.}
\label{tab:stopmain}
\begin{tabular}{@{}lrrrr@{}}
\toprule
 & \multicolumn{2}{c}{R1-Distill-Qwen 1.5B} & \multicolumn{2}{c}{R1-Distill-Qwen 7B}\\
\cmidrule(lr){2-3}\cmidrule(lr){4-5}
Policy & Accuracy & Tokens & Accuracy & Tokens\\
\midrule
Full budget & $\fullAcc$ & $\fullTokens$ & $\sevBfullAcc$ & $\sevBfullTokens$\\
Oracle stopping (gold labels) & $\orcAcc$ & $\orcTokens$ & $\sevBorcAcc$ & $\sevBorcTokens$\\
Drift alarm (ours) & $\alarmAcc$ & $\alarmTokens$ & $\sevBalarmAcc$ & $\sevBalarmTokens$\\
Checkpoint vote & $\voteAcc$ & $\voteTokens$ & $\sevBvoteAcc$ & $\sevBvoteTokens$\\
Answer agreement & $\agreeAcc$ & $\agreeTokens$ & $\sevBagreeAcc$ & $\sevBagreeTokens$\\
\bottomrule
\end{tabular}
\end{table}

None of the $\blNimpl$ implementable configurations at 1.5B ($\sevBblNimpl$ at
7B), which include fixed truncation and the answer-convergence,
self-consistency and adaptive-sampling rules from the literature, improves
accuracy at lower charged token cost. Rules that save tokens change accuracy by between $\blSaveMild$
and $\blSaveWorst$ points at 1.5B. Our drift alarm, which stops when a learned
critic predicts accumulated erosion, does not improve on them either
(App.~\ref{app:diagnostics}). The census in Sec.~\ref{sec:exp1} explains why:
the states worth stopping at are few and concentrated in particular problems,
so even perfect detection would add about two points.

\section{Related Work}
\label{sec:related}

\paragraph{Overthinking and adaptive compute.} Work on overthinking studies
wasted computation on easy problems \citep{chen2024overthinking,sui2025stopoverthinking},
budget-dependent accuracy \citep{zhou2026morethinking}, answer abandonment
\citep{wei2025evolution} and agentic over-reasoning \citep{cuadron2025danger}.
Early-exit methods use answer convergence, redundancy detection, attention
probes, cost-aware stopping and adaptive sampling
\citep{zhang2025reasoning,liu2025answerconvergence,mao2025escot,%
kadavath2022know,sun2026refrain,min2026puma,li2026asag,dong2026learnstop,%
zabounidis2026reforc,aggarwal2023adaptive,huang2025selfcalibration}. Our
decomposition shows that survivor-averaged budget curves overstate within-chain
harm, and our headroom measurement bounds what such methods can add in
accuracy.

\paragraph{Value estimation and optimal stopping.} Process reward models and
value probes are often trained on Monte Carlo rollout labels: the fraction of
continuations from a prefix that end correct
\citep{wang2024mathshepherd,lightman2024verify}. That fraction estimates the
verdict. Deciding whether to stop also needs the commitment from the same
prefix, the stop-now value of optimal stopping \citep{chow1971great}.

\paragraph{Martingale views of reasoning.} \citet{li2026foresight} use the Doob
decomposition and stopping theory for path selection, \citet{kim2026drift}
study coherence under answer resampling, \citet{farmer2026alphalaw}
characterize elicited probability transformations, and
\citet{he2025martingale} test self-reported belief revisions against a
martingale rationality criterion. We separate the forecast of the final outcome
from the quality of the answer available now, and connect their gap to the
value of continuing.

\section{Discussion and Limitations}
\label{sec:conclusion}

Deciding whether more reasoning is worth it requires comparing the answer
available now with the answer continuing would produce, on the same chain.
Final-correctness forecasts and survivor-averaged budget curves each miss one
side of this comparison. Measured directly, continuing rarely loses a correct
answer in the models we study; the losses concentrate in a few problems, and
the declining curves reported in the literature come largely from which chains
remain. The same comparison applies to other ways of spending computation, such
as self-revision rounds or multi-agent exchanges (App.~\ref{app:beyond}).

\paragraph{Limitations.} Our experiments cover mathematical reasoning and
models up to 8B parameters. Commitment is defined by a specific stop-and-answer
procedure; the robustness checks in App.~\ref{app:expdetails} vary it. The
training objectives of App.~\ref{app:objective} are tested only on synthetic
data.

\clearpage
\subsubsection*{Reproducibility statement}
All experiments run on single consumer GPUs (12 and 16\,GB); models and
datasets are public. Seeds (0 for the primary audits; the resampling and ablation studies
draw from disjoint seed streams recorded in the released code), prompts and
forcing strings, splits,
hyperparameters, and the pre-committed gates are in
App.~\ref{app:expdetails}. The matched follow-up and refit variant have separate
frozen settings, raw outputs, and checks (Apps.~\ref{app:matched}--\ref{app:refits});
their seed streams start at 20260907. Analyses added during review
(App.~\ref{app:rebuttal}) have their own dated plan, fixed before
generation; their seed streams start at 20261106. Quantitative tables are checked against
result files. Internal dated plans are distinguished from external registration.
The new follow-up code and results accompany the primary audit artifacts.

\subsubsection*{AI Use Statement}
An AI coding assistant was used throughout this project: for drafting and
editing prose, writing and debugging experiment and analysis code, scaffolding
statistical analyses, and assisting with mathematical reformulation and proof
checking. The authors take responsibility for the final text, theoretical
claims, proofs, code, analyses, and reported results, including any errors.

\bibliographystyle{iclr2027_conference}
\bibliography{references,main2_related}

\appendix

\section{Notation Glossary}
\label{app:glossary}

\begin{center}
\small
\begin{tabular}{@{}p{0.22\textwidth}p{0.74\textwidth}@{}}
\toprule
$h_t$, $\F_t$, $z_t$ & complete reasoning state; its filtration; observed tap $z_t=f(h_t)$\\
$\pi_t$, $\hat\pi_t$ & commitment belief (answer-if-stopped-now correct); its probe\\
$\Phi_t$, $\Phi(h)$, $\hat\Phi$ & verdict $\E[Y\mid\F_t]$; its state form, equal to $\Phi_t$ under Assumption~1 (App.~\ref{app:operator}); its probe\\
$\Delta_t=b_t+\eta_t$ & belief increment = predictable drift $+$ innovation\\
$m_t=\E[\Delta_t\mid z_t,t]$ & tap-conditional drift (what a critic can see)\\
$B_t$, $\hat B_t$ & accumulated drift (Doob compensator); its estimate\\
$g_\phi(h_t,t)$ & drift critic (sensor); $\tilde\pi_t=\hat\pi_t-\hat B_t$ compensated belief\\
Signal A / B & forced-answer correctness (probe-independent) / probe readout\\
\ow{}, \sco{} & correct at a checkpoint $\le2048$ then wrong / then right at the end\\
\bottomrule
\end{tabular}
\end{center}

\section{Full Audit Matrix}
\label{app:auditmatrix}

\begin{table}[h]
\footnotesize
\centering
\caption{Full pre-registered gate and post-hoc robustness matrix; quantities
1.5B\,/\,7B. The pre-registered gates were fixed before the corresponding
real-data runs; post-hoc analyses are distinguished below and in the text.}
\label{tab:auditfull}
\begin{tabular}{@{}p{0.28\textwidth}p{0.68\textwidth}@{}}
\toprule
\textbf{Pre-registered} & \textbf{outcome}\\
\midrule
Gate 0: apparatus calibration & passed (slope $\eZeroSlopeMid$, $R^2\,\eZeroRTwoMid$; matched floor $\eZeroFloorMatched$)\\
Gate 1(a): $\ge30$ \ow{} chains & failed ($10/367$) $\to$ committed 7B escalation: $9/262$\\
Gate 2: recovery at some $\kappa$ & not met: no tested alarm recovers accuracy; this does not exclude local adverse drift\\
Verdict rule (viii) & not triggered: verdict AUC $\verdictAUCthree$/$\sevBverdictAUCthree$ does not clear $0.75$ (App.~\ref{app:expdetails})\\
\midrule
\textbf{Robustness} & \textbf{outcome}\\
\midrule
Forcing operator & of $\fvPairs$ truncations, $\fvNoAnsPct\%$ yield no natural answer; on the rest, agreement $\fvCorrAgree\%$; greedy vs.\ sampled $\dmDelta$ pts $\dmCI$\\
Checkpoint grid & density inflates \ow{}: $\gCoarseOW$ (1024) $=\gMidOW$ (512) $\to\gDenseOW$ (256) of $\gDenseN$\\
Label definitions & prevalence $1$--$4\%$ under all final-outcome variants\\
Critic capacity & linear critic keeps the signal at 1.5B ($R^2$ $\linRtwo$ vs.\ $\mlpRtwo$)\\
Alarm threshold & no percentile in $\{50..99\}$ beats full budget (95th: $\consKapDelta$ pts)\\
Tap richness & ordering holds only at the 1.5B primary tap; unresolved overall\\
Self-consistency & vote curves inherit survivorship ($\scVoteWithin$ within / $\scVoteComp$ comp.)\\
Third model & Qwen3-1.7B (non-R1-distilled, Qwen lineage): \ow{} $\qwenPct\%$ $\qwenCI$; survivorship reproduces\\
Other base families & R1-Distill-Llama 8B: \ow{} $\llamaPct\%$ $\llamaCI$, within $\llamaWithin$ / comp.\ $\llamaComp$; Phi-4-mini: $\phiPct\%$ $\phiCI$, $\phiWithin$ / $\phiComp$\\
Decoding temperature & at $0.9$: \ow{} $\tmpAltRate\%$ (vs.\ $\tmpDefRate\%$); within-cohort still positive\\
Synthetic sweep & median recovery $\sweepRecMed\%$ over 19 configs; failure mode: belief saturation\\
\bottomrule
\end{tabular}
\end{table}

\section{Extended Related Work}
\label{app:relext}

\paragraph{Probes, values, estimation.} Hidden states encode answer correctness \citep{zhang2025reasoning} and
linearly represent belief-state geometry \citep{shai2024belief}---the license
for our readouts: commitments can drift, whereas exact verdicts on nested
histories cannot; lossy instantaneous taps and fitted probes need not inherit
that null (App.~\ref{app:observation}). Process reward models and MC/TD
variants estimate step values
\citep{lightman2024verify,wang2024mathshepherd,zhang2025tdrm}; ours is
orthogonal.

The training objective is adversarial conditional-moment estimation
\citep{dikkala2020minimax,lewis2018adversarial} in the predictability-minimization
family \citep{schmidhuber1992predictability,ganin2016dann,zhang2018adversarialdebias};
post-hoc self-revision alone is unreliable
\citep{huang2024selfcorrect,madaan2023selfrefine}.

\section{Proofs}
\label{app:proofs}

The results of Secs.~\ref{sec:setup}--\ref{sec:doob} use the following
assumption on the reasoning process.

\begin{assumption}\label{ass:markov}
(A1) The sampling noise $\xi_t$ is i.i.d.\ and independent of $\F_t$, so $(h_t)$
is a time-homogeneous Markov chain with operator
$(Pf)(h)=\E[f(h_{t+1})\mid h_t=h]$; and terminal correctness $Y$ is a function
of the state in which the chain absorbs.
(A2) The readout is a measurable function of $h_t$ alone (lossy taps:
App.~\ref{app:diagnostics}); a critic receiving the step index augments the state to
$(h_t,t)$, for which all results hold.
\end{assumption}

\begin{theorem}[Ballistic bias, diffusive variance]
\label{thm:ballistic}
If $|b_t|\le\beta$ and $|\eta_t|\le c$ a.s., then $\pi_T-\pi_0=B_T+(M_T-M_0)$ with
$B_T=\sum_{t<T}b_t$, $|B_T|\le\beta T$, and $(M_t)$ a martingale with increments bounded
by $c$, so $\Pr(|M_T-M_0|\ge\varepsilon)\le2\exp(-\varepsilon^2/2Tc^2)$
(Azuma--Hoeffding). If, for a fixed sign $s\in\{-1,1\}$,
$s b_t\ge\beta_0>0$ a.s.\ for every $t<T$, then with probability at least
$1-\alpha$, for $0<\alpha<1$,
\[
 s(\pi_T-\pi_0)\ge\beta_0T-c\sqrt{2T\log(1/\alpha)}.
\]
This is a finite-horizon statement. On the probability scale the persistence
assumption entails $\beta_0T\le1$ in expectation; it is not an asymptotic
growth law for bounded beliefs. Log-odds drift requires its own decomposition.
\end{theorem}

\paragraph{Theorem~\ref{thm:ballistic}.}
Summing \eqref{eq:split}, $M_t:=\pi_t-\sum_{s<t}b_s$ has
$\E[M_{t+1}-M_t\mid\F_t]=\E[\eta_t\mid\F_t]=0$ and increments bounded by $c$;
Azuma--Hoeffding gives the tail bound. $|B_T|\le\beta T$ is the triangle inequality.
The one-sided inequality applied to $s(M_T-M_0)$ gives the displayed signed
bound. Taking expectations under $s b_t\ge\beta_0$ gives
$\beta_0T\le\E[s(\pi_T-\pi_0)]\le1$. \qed

\paragraph{Ensembling: innovation and shared drift.}\label{cor:onlybias}
Averaging $k$ independent belief processes reduces the standard deviation of
their averaged innovation by a factor of $\sqrt{k}$ under identical
variances, but leaves common expected drift intact. This concerns belief
averaging, not nonlinear voting over terminal answers; vote-curve
survivorship is audited in App.~\ref{app:audit2}.

\begin{theorem}[Doob]
\label{thm:doob}
Any integrable adapted $(\pi_t)$ decomposes uniquely as $\pi_t=M_t+B_t$ with $(M_t)$ a
martingale and $(B_t)$ predictable, $B_0=0$; necessarily
$B_t=\sum_{s<t}\E[\Delta_s\mid\F_s]$.
\end{theorem}

Uniqueness (App.~\ref{app:proofs}) makes the construction canonical: with
``systematic drift'' meaning the predictable part, there is exactly one such
object.

\begin{definition}[Commitment conservation]
\label{def:calibrated}
The commitment is \emph{conserved} (dynamically calibrated) iff
$B_t\equiv0$, equivalently $\E[\pi_{t+1}\mid\F_t]=\pi_t$.
This is a dynamics condition, not standard probabilistic calibration, a
rationality axiom, or a correctness guarantee: even a constant wrong
readout satisfies it, whereas useful computation can increase commitment.
\end{definition}
At certainty, $b_t=\E[\pi_{t+1}\mid\F_t]-1\le0$ on $\{\pi_t=1\}$,
strictly if a decrease has positive conditional probability. A conserved
bounded commitment cannot leave $1$. A sampled correct answer need not
imply certainty; deterministic elicitation at a fully specified state does
(App.~\ref{app:proofs}).

\paragraph{Theorem~\ref{thm:doob}.}
Existence by the displayed construction. Uniqueness: if $\pi=M+B=M'+B'$ with both
decompositions as stated, $N:=M-M'=B'-B$ is a predictable martingale with $N_0=0$, so
$N_{t+1}=\E[N_{t+1}\mid\F_t]=N_t=\dots=0$. \qed

\paragraph{Proposition~\ref{prop:verdict}.}
(i) $\E[\Phi_{t+1}\mid\F_t]=\E[\E[Y\mid\F_{t+1}]\mid\F_t]=\E[Y\mid\F_t]=\Phi_t$
by the tower property, since $\F_t\subseteq\F_{t+1}$; $Y$ is bounded, so all
conditional expectations exist. No further hypothesis is used---in particular
none on the dynamics, the tap, or the stopping time---so a chain that overthinks
has a conserved verdict exactly as one that does not. Its Doob decomposition
(Thm.~\ref{thm:doob}) is $\Phi_t=\Phi_t+0$, so $B\equiv0$ by uniqueness.
(ii) For $s\ge t$, $\E[\pi_{s+1}-\pi_s\mid\F_t]=\E[b_s\mid\F_t]$; summing
from $s=t$ to $S$ and letting $S\to\infty$, the commitment side converges by
conditional dominated convergence (the chain is frozen after $\tau$, so
$\pi_S\to\pi_\tau$ pointwise; $|\pi|\le1$ and $\pi_\tau=Y$ give
$\E[\lim_S\pi_S\mid\F_t]=\E[Y\mid\F_t]=\Phi_t$) and the drift side is
dominated by $(\tau-t)_+$ (as $b_s=0$ for $s\ge\tau$ and $|b_s|\le1$), integrable
because $\E[\tau]<\infty$. This gives
$\Phi_t-\pi_t=\E[\sum_{s\ge t}b_s\mid\F_t]$, i.e.\ \eqref{eq:gap}. If
$\pi_t=\Phi_t$ a.s.\ for all $t$ then $(\pi_t)$ is a martingale, so
$B\equiv0$ by uniqueness; conversely $B\equiv0$ makes the right side vanish.
Writing $\Phi_t=\Phi(h_t)$ for a measurable $\Phi$ requires additionally that
$h_t$ be sufficient for $\F_t$ with respect to $Y$ (Assumption~\ref{ass:markov}),
which is used nowhere above and is an assumption about the \emph{observation},
not the process. \qed

\begin{theorem}[Convergence and budget-invariance]
\label{thm:budget}
A calibrated belief ($B_t\equiv0$) converges a.s.\ and in $L^1$ and, for every
stopping time $\tau\le T$, satisfies
$\E[\pi_\tau\mid\F_0]=\pi_0$. Thus its expected conclusion is
budget-invariant; a biased chain instead adds $\E[B_\tau\mid\F_0]$.
\end{theorem}
This concerns the mean, not the full stopped distribution; interior beliefs
can decrease on individual paths. Calibration also yields a polarization
bound (Thm.~\ref{thm:steps}); adverse drift produces an interior optimal budget
in the synthetic model. Proofs are in App.~\ref{app:proofs}.

\paragraph{Theorem~\ref{thm:budget}.}
Convergence: Doob's martingale convergence theorem for bounded martingales; $L^1$ by
uniform integrability. Budget-invariance: optional stopping for bounded martingales
($\tau\le T$). For the biased case apply optional stopping to $M_t=\pi_t-B_t$. \qed

\paragraph{Certainty at the boundary.}
If $0\le\pi_t\le1$, then on $\{\pi_t=1\}$ we have
$b_t=-\E[1-\pi_{t+1}\mid\F_t]\le0$. The inequality is strict wherever
$\Pr(\pi_{t+1}<1\mid\F_t)>0$. If $b_t=0$, the nonnegative variable
$1-\pi_{t+1}$ has conditional expectation zero and therefore vanishes a.s.
Iteration makes $1$ absorbing for a bounded martingale. The same argument
applies to $0$. Thus a single stochastic correct elicitation is not proof of
certainty, but an exact binary correctness readout cannot undergo a
positive-probability correct-to-wrong transition with zero adverse drift at
every preceding correct state. With lossy taps or learned probabilities,
this implication cannot be inferred from one observed label alone. \qed

\begin{theorem}[Polarization / step count]
\label{thm:steps}
Assume calibration, $0<\sigma^2\le1$, and relative innovation non-degeneracy,
$\E[\eta_t^2\mid\F_t]\ge\sigma^2\,\pi_t(1-\pi_t)$ for all $t$ (the right-hand side
vanishes at $\pi_t\in\{0,1\}$, so absorbing endpoints remain admissible). Then
$\E[\pi_T(1-\pi_T)]\le(1-\sigma^2)^T/4$. For $0<\varepsilon\le1/2$ and $0<\delta<1$, after
$T=O\!\big(\sigma^{-2}\log(1/\varepsilon\delta)\big)$ steps the belief has exited to
within $\varepsilon$ of a definitive verdict with probability $\ge1-\delta$, and
bounded threshold stopping preserves the mean by Thm.~\ref{thm:budget}. More information per step (larger
$\sigma$) means fewer steps; the dependence on the target is logarithmic.
\end{theorem}

\paragraph{Theorem~\ref{thm:steps}.}
Let $v_t=\pi_t(1-\pi_t)$. Calibration gives
$\E[v_{t+1}\mid\F_t]=v_t-\E[\eta_t^2\mid\F_t]\le v_t(1-\sigma^2)$ for every $t$, using
$\E[\eta_t^2\mid\F_t]\ge\sigma^2 v_t$. Iterating, $\E[v_T]\le(1-\sigma^2)^T/4$. On
$\{\pi_T\in(\varepsilon,1-\varepsilon)\}$, $v_T\ge\varepsilon(1-\varepsilon)\ge
\varepsilon/2$, so Markov's inequality gives
$\Pr(\pi_T\in(\varepsilon,1-\varepsilon))\le(1-\sigma^2)^T/(2\varepsilon)\le\delta$ for
$T\ge\sigma^{-2}\log(1/(2\varepsilon\delta))$. Validity of threshold stopping is
Thm.~\ref{thm:budget}. \qed

\paragraph{Accuracy is not posterior polarization.}
For a calibrated posterior $p_t$ on a fixed binary hypothesis, the Bayes
decision accuracy is $\E[\max\{p_t,1-p_t\}]$, which can increase while
$\E[p_t]$ is constant. For an answer-if-stopped-now success probability
$\pi_t$, expected accuracy is instead $\E[\pi_t]$. These are different
readouts. In particular, a martingale decomposition alone does not justify an
accuracy expansion of the form $a+c\sqrt T-dT$, or an interior optimal budget.
The interior peak in Sec.~\ref{sec:exp0} is a property of the specified
synthetic decision model, measured numerically. Equation~\eqref{eq:advantage}
is the corresponding exact statement for operational commitment accuracy.

\paragraph{Proposition~\ref{prop:quadratic}.}
$\E[\Delta_t^2]=\E[(b_t+\eta_t)^2]=\E[b_t^2]+\E[\eta_t^2]$ since the cross term
vanishes by the tower property ($b_t$ is $\F_t$-measurable). A penalty
$\rho(\Delta_t)$ with $\rho$ locally quadratic charges both terms at the same
second-order rate. Consecutive-step KL between readout distributions is, to second
order, at interior probabilities, the Fisher-weighted squared increment
$\Delta_t^2/(2\pi_t(1-\pi_t))$; the weight
is $\F_t$-measurable, so $\E[w\Delta_t^2]=\E[w\,b_t^2]+\E[w\,\eta_t^2]$ with the same
weight on both components. This is a local expansion away from the endpoints,
not a global comparison between KL and squared increments. \qed

\paragraph{Lemma~\ref{lem:var}.}
$2\E[g\Delta]-\E[g^2]=2\E[g\,m]-\E[g^2]=\E[m^2]-\E[(g-m)^2]$ using the tower property
($\E[g(h_t)\Delta_t]=\E[g(h_t)m_t]$), maximized at $g=m$ over any class containing
$m$; for general $\G$ the display equals $\E[m^2]-\E[(g-m)^2]$, whose supremum over
$\G$ is $\E[m^2]-\inf_{g\in\G}\E[(g-m)^2]$. \qed

\paragraph{Theorem~\ref{thm:correct}.}
(i) Sum Lemma~\ref{lem:var} over $t$ (the critic receives the step index, so the sum is
attained term-wise); under Assumption~\ref{ass:markov}, $m_t=b_t$. (ii) If $m_t=0$ then
$\E[g\Delta_t]=\E[g\,m_t]=0$ for every $g\in\G$, so the inner objective is
$-\E[g^2]\le0$, maximized at $g\equiv0$ with value $0$. \qed

\begin{proposition}[Graceful degradation]
\label{prop:graceful}
Under Assumption~\ref{ass:markov}, if $|g_\phi(h_t,t)-m_t|\le\delta$, then for
any stopping time $\tau$ with $\E[\tau]<\infty$,
$\big|\E[\tilde\pi_\tau-\tilde\pi_0]\big|\le\delta\,\E[\tau]$, and conditionally
$\big|\E[\tilde\pi_\tau\mid\F_0]-\tilde\pi_0\big|\le\delta\,\E[\tau\mid\F_0]$.
\end{proposition}

\paragraph{Proof.}
$\tilde\pi_{t+1}-\tilde\pi_t=\Delta_t-g_\phi(h_t,t)$ has conditional mean
$b_t-g_\phi(h_t,t)=m_t-g_\phi(h_t,t)\in[-\delta,\delta]$ under
Assumption~\ref{ass:markov}. Write $\tilde\pi_t=N_t+D_t$ with $(N_t)$ a
martingale ($N_0=\tilde\pi_0$, increments equal to
$\Delta_t-b_t$ and bounded by $2$) and $(D_t)$ predictable with
$D_0=0$ and $|D_{t+1}-D_t|\le\delta$. Optional stopping for
bounded-increment martingales at $\tau$ with $\E[\tau]<\infty$ gives
$\E[N_\tau]=\E[N_0]=\E[\tilde\pi_0]$, and $|D_\tau|\le\delta\tau$ gives
$|\E[\tilde\pi_\tau-\tilde\pi_0]|=|\E[D_\tau]|\le\delta\,\E[\tau]$; conditioning
on $\F_0$ throughout gives the second form. \qed

\section{The Verdict in Operator Form}
\label{app:operator}

\paragraph{Harmonic reference and accumulated drift.}
The common terminal boundary in Prop.~\ref{prop:verdict}(ii) makes the gap
equal to accumulated \emph{remaining} expected change. In Markov form, the
same relationship becomes a Poisson equation. Under
Assumption~\ref{ass:markov}, the exact value function
$\Phi(h)=\Pr(Y=1\mid h)$ satisfies $P\Phi=\Phi$: it is harmonic for the
fixed continuation policy. Therefore
\begin{equation}
 b=(P-I)\pi=(P-I)(\pi-\Phi).
 \label{eq:poisson}
\end{equation}
In a finite absorbing chain, $\Phi$ is the unique bounded harmonic extension
of terminal correctness and $\pi-\Phi=-(I-Q)^{-1}b$, with $Q$ the
transient-state matrix (see below). Thus the gap weights signed
local drift by expected future state visits; positive and negative changes
can offset. In particular, $|\pi(h)-\Phi(h)|\le\|b\|_\infty\E_h[\tau]$.
The expected occupation of future states determines how much each local
drift contributes. A harmful step followed by sufficiently beneficial ones
can still leave positive continuation value. The Poisson equation thus
connects stepwise answer dynamics to the value of the entire remaining path.

Let $(Pf)(h)=\E[f(h_{t+1})\mid h_t=h]$ (Assumption~\ref{ass:markov}) and $L=P-I$.

\begin{proposition}[Operator form]
\label{prop:operator}
(a) $b_t=(L\pi)(h_t)$: drift is the generator applied to the readout. (b) The chain is
calibrated iff $P\pi=\pi$, i.e.\ the readout is \emph{harmonic}: $\pi\in\ker L$. (c)
Assume the terminal structure is \emph{absorbing}: from every interior state the
boundary (an emitted answer, carrying $g_\partial=\mathbf 1\{\text{correct}\}$) is
reached with probability one. Then $\Phi$ is the unique \emph{bounded} solution of the
discrete Dirichlet problem $L\Phi=0$ with boundary data $g_\partial$, and
$\Phi(h)=\E_h[g_\partial(h_\tau)]=V(h)$: the verdict is the harmonic extension of
terminal correctness, i.e.\ the exact value function. Under
Assumption~\ref{ass:markov} this state function agrees with the
filtration-conditional verdict of Prop.~\ref{prop:verdict},
$\Phi(h_t)=\E[Y\mid\F_t]=\Phi_t$ a.s., which is what licenses writing $\Phi$ as
a function of the state at all. (d) $\sum_t\E[b_t^2]
=\lVert(P-I)\pi\rVert^2_{L^2(\mu)}$, where
$\mu(A)=\E\big[\sum_{t<\tau}\mathbf 1\{h_t\in A\}\big]$ is the
\emph{unnormalized} pre-absorption occupancy measure (no horizon factor is
needed with this normalization): the
training objective of Sec.~\ref{sec:objective} is a squared generator residual,
minimized exactly on harmonic readouts on the occupied states. It is not the
usual Dirichlet form $\langle\pi,(I-P)\pi\rangle$.
\end{proposition}

\begin{proof}
(a), (b) are definitional; (d) sums pointwise. For (c), define
$\Phi(h)=\E_h[g_\partial(h_\tau)]$. Almost-sure absorption makes this bounded
candidate well-defined; the Markov property gives $P\Phi=\Phi$ in the interior
and its boundary values are $g_\partial$. Uniqueness holds because if $u$ is bounded, harmonic, and zero
on the boundary, then $u(h)=\E_h[u(h_{t\wedge\tau})]\to0$ by dominated convergence under
a.s.\ absorption. Without absorption uniqueness genuinely fails:
$u(h)=\Pr_h(\tau=\infty)$ is bounded, harmonic, and vanishes on the boundary.
\end{proof}

\paragraph{Fundamental-matrix form (finite case).} If the interior state space is finite
(e.g.\ a finite discretization, or the state augmented with a step index under a hard
budget $T_{\max}$, which makes the interior block nilpotent) and the chain is absorbing,
write $P=\begin{psmallmatrix}Q&R\\0&I\end{psmallmatrix}$. Then $Q^t\to0$, hence
$\rho(Q)<1$, $I-Q$ is invertible, and
\begin{equation}
  \Phi_{\mathrm{int}}=(I-Q)^{-1}R\,g_\partial,\qquad
  (I-Q)^{-1}=\sum_{t\ge0}Q^t
  \label{eq:green}
\end{equation}
\citep[Thm.~3.2.1]{kemeny1976finite}: $N=(I-Q)^{-1}$ is the fundamental matrix (the
Green's operator of $L$), $N_{hh'}$ the expected visits to $h'$ before absorption, and
$NR$ the absorption-distribution map. On infinite state spaces sub-stochasticity does
\emph{not} imply $\rho(Q)<1$; the probabilistic representation of
Prop.~\ref{prop:operator}(c) is the correct general statement, and \eqref{eq:green}
has a probabilistic interpretation through visit counts. Almost-sure absorption
alone does not ensure finite expected duration or a bounded inverse on an
infinite state space.

\paragraph{Poisson representation and horizon sensitivity.}
Writing $d=\pi-\Phi$ on the transient states, the common terminal boundary gives
$(I-Q)d=-b$, hence $d=-Nb$. Consequently
\[
 |d(h)|\le\E_h\!\left[\sum_{t<\tau}|b(h_t)|\right]
 \le\|b\|_\infty\E_h[\tau].
\]
Expected remaining duration therefore controls how local drift translates
into a commitment--verdict gap. In the finite absorbing case $Q^t\to0$;
if $Q$ is diagonalizable, convergence is bounded by a constant times
$\rho(Q)^t$. Diagonalizability of an arbitrary Markov operator alone is
insufficient: other unit-modulus eigenvalues can prevent convergence.

\paragraph{On log-odds ``closed forms.''} The compensated belief
$\tilde\pi_t=\pi_t-\hat B_t$ estimates the \emph{martingale component} $M_t$
(Thm.~\ref{thm:doob}); it differs from the verdict $\Phi_t$ by the expected
\emph{total} accumulated drift $\E[B_\infty\mid\F_t]$, while the raw belief differs
from $\Phi_t$ by expected \emph{future} drift alone \eqref{eq:gap}. No identity of the form $\Phi=\sigma(\langle u_A,h\rangle-\sum_{t\ge s}b_t)$
holds: expectations do not commute with $\sigma$, and probability-scale drift is not a
log-odds increment. If a log-odds compensation is wanted, define the log-odds drift
$\tilde b_t=\E[L_{t+1}-L_t\mid\F_t]$ for $L_t=\mathrm{logit}\,\pi_t$ and compensate
inside the logit; we use the probability scale throughout.

\section{The Variance-Blind Objective: Formal Statements}
\label{app:objective}

\paragraph{Selective improvement.}
The sign of $b_t$ distinguishes improvement from erosion, and the sum of
future signed drifts determines the value of the remaining computation.
This makes the choice of a training target consequential.
A penalty on all answer movement,
$\E[\Delta_t^2]=\E[b_t^2]+\E[\eta_t^2]$, suppresses both predictable change
and stochastic exploration. A penalty on $b_t^2$ isolates predictable
change but also charges beneficial improvement. A one-sided target,
$\E[(b_t^-)^2]$ with $b_t^-=\max\{-b_t,0\}$, instead isolates expected erosion.
The rest of this appendix derives the corresponding variational objectives.
The mathematical distinction is useful for process supervision: rewarding
better answers and preserving available correct answers require a signed
quality signal.

This appendix states and discusses the results summarized in
Sec.~\ref{sec:objective}. In the critic formulas, $h_t$ denotes the available
input (the tap $z_t$ in App.~\ref{app:diagnostics}), with time included when supplied.
Equality $m_t=b_t$ requires a sufficient input; otherwise the objectives target
only the predictable component visible to that input.

\begin{proposition}[One-sided erosion objective]
\label{prop:onesided}
For a square-integrable increment $\Delta_t$, let
$m_t=\E[\Delta_t\mid z_t,t]$ and $m_t^-=\max\{-m_t,0\}$. Then
\[
 \sup_{g(z_t,t)\le0}\{2\E[g\Delta_t]-\E[g^2]\}
 =\E[(m_t^-)^2],\qquad g^\star=\min\{m_t,0\},
\]
where the supremum is over square-integrable, measurable functions. The value
is zero iff $m_t\ge0$ a.s.; thus it penalizes adverse predictable movement
without directly penalizing beneficial predictable movement or residual variance.
\end{proposition}
\begin{proof}
Conditioning gives $2\E[g\Delta_t]-\E[g^2]=\E[2gm_t-g^2]$.
Pointwise, the concave quadratic is maximized over $g\le0$ at
$\min\{m_t,0\}$, with value $(m_t^-)^2$. This maximizer is square-integrable
by conditional Jensen. A restricted critic class containing zero has a
nonnegative supremum no larger than this value, with equality when it contains
the maximizer.
\end{proof}
This is a derived alternative to the symmetric objective, not a reported
training experiment. Under a sufficient tap, its zero set is the
submartingale condition $b_t\ge0$, which by Cor.~\ref{cor:advantage} excludes
an accuracy-improving stopping rule under the same protocol. Estimator error,
representation shift, and task loss still matter during optimization.

\begin{proposition}[Readout error propagates into measured drift]
\label{prop:readouterror}
Let $\hat\pi_t=\pi_t+e_t$ be adapted with square-integrable errors, and define
$\hat b_t=\E[\hat\pi_{t+1}-\hat\pi_t\mid\F_t]$. Then
\[
 \hat b_t-b_t=\E[e_{t+1}\mid\F_t]-e_t,\qquad
 \|\hat b_t-b_t\|_2\le\|e_{t+1}\|_2+\|e_t\|_2.
\]
If $|e_t|\le\epsilon$ a.s.\ at every step, the pointwise drift error is at
most $2\epsilon$. The same bound holds after projection onto a tap.
\end{proposition}
\begin{proof}
Subtract the conditional increment definitions and apply conditional
expectation contraction in $L^2$, followed by the triangle inequality.
The uniform bound follows directly; another conditional expectation is also
a contraction.
\end{proof}
Neither AUC nor binned ECE supplies such a uniform error bound. Consequently,
predictable increments of a learned probe certify predictability of that
probe, not automatically of the underlying behavioral success probability.

\paragraph{The objective, as used in the main text.} Penalizing the variational
value of Sec.~\ref{sec:objective} gives the label-free, variance-blind objective
\begin{equation*}
\min_\theta\;\Lt(\theta)+\lambda\max_\phi\sum_t
\big(2\E[g_\phi(h_t,t)\Delta_t]-\E[g_\phi(h_t,t)^2]\big).
\end{equation*}
With a fixed readout and a critic class containing $m_t$, the inner maximum is
exactly $\sum_t\E[m_t^2]$: it vanishes iff the chain is state-calibrated, while
admissible zero-mean innovations incur no direct penalty.

\begin{proposition}[Locally quadratic penalties charge both components]
\label{prop:quadratic}
Since $b_t$ is $\F_t$-measurable, $\E[\Delta_t^2]=\E[b_t^2]+\E[\eta_t^2]$. Any penalty
that is locally quadratic in the per-step belief change---squared-increment
regularizers and, to second order, KL-type proximity penalties between consecutive \emph{readout} distributions---charges innovation on the same footing as drift.
It therefore penalizes useful unpredictable updates as well as predictable movement;
the effect of optimization on either component depends on the model and task loss.
\end{proposition}

We penalize only the predictable part. As a conditional expectation, it admits a
variational characterization---the classical $L^2$ projection, used here as an
\emph{adversarial conditional-moment} estimator
\citep{dikkala2020minimax,lewis2018adversarial}.

\begin{lemma}[Variational identity]
\label{lem:var}
Let $m_t=\E[\Delta_t\mid h_t]$ and let $\G$ be a function class containing $m_t$. Then
\begin{equation}
  \E[m_t^2]=\max_{g\in\G}\big\{2\,\E[g(h_t)\,\Delta_t]-\E[g(h_t)^2]\big\},
  \label{eq:var}
\end{equation}
attained at $g^\star=m_t$. For a capacity-limited $\G$, the supremum equals
$\E[m_t^2]-\inf_{g\in\G}\E[(g-m_t)^2]$, the representable share of the drift energy:
guarantees hold against drift representable by the critic class.
\end{lemma}

\begin{definition}[Drift-penalized training]
\label{def:training}
With a frozen readout and a critic $g_\phi(h_t,t)$ (state and step index),
\begin{equation}
  \min_\theta\;\Lt(\theta)\;+\;\lambda\,\max_\phi\sum_t
  \big(2\,\E[g_\phi(h_t,t)\,\Delta_t]-\E[g_\phi(h_t,t)^2]\big).
  \label{eq:train}
\end{equation}
The critic forecasts the next belief move; the reasoner is trained to make its own
updates unforecastable. The penalty uses \emph{no gold labels}.
\end{definition}

\begin{theorem}[Identification of drift energy and variance-blindness]
\label{thm:correct}
(i) Provided $\G$ contains the map $(h,t)\mapsto m_t(h)$, the inner maximum of
\eqref{eq:train} equals $\sum_t\E[m_t^2]$; it is zero iff the
chain is state-calibrated ($m_t\equiv0$), and under Assumption~\ref{ass:markov} equals
$\sum_t\E[b_t^2]$ with zero iff $B_t\equiv0$. (ii) \emph{Variance-blindness}: if
$m_t=0$ with $\eta_t$ arbitrary, the optimal critic is $g\equiv0$ and the penalty
vanishes. The objective does not directly penalize zero-mean innovations;
this orthogonality does not guarantee that optimizing a shared model preserves
accuracy, information acquisition, or innovation variance.
\end{theorem}

Algorithm~\ref{alg:train} (App.~\ref{app:deploy}) instantiates \eqref{eq:train} with a
two-timescale update. Freezing the head fixes its weights but does not fix its
input representation or calibration. Equation~\eqref{eq:var} descends from
predictability minimization and adversarial conditional-moment estimation (lineage in
Sec.~\ref{sec:related}); what is new is the semantics---the value held fixed, the
\emph{dynamics} regularized, the variance-blindness guarantee (ii), and the critic's
second job at inference (Sec.~\ref{sec:control}). Sec.~\ref{sec:exp0} validates the
mechanism where ground truth is available; fine-tuning a full reasoning model with
\eqref{eq:train} is future work (App.~\ref{app:deploy}).

\section{Deployment Notes and Future Training Work}
\label{app:deploy}

\paragraph{Serving (Level 1).} One critic forward pass per step; maintain
$\hat B_t=\sum_{s<t}g_\phi(h_s,s)$; alarm when $-\hat B_t>\kappa$; answer from the last
safe checkpoint (or force an answer at the trigger point). The compensated belief
$\tilde\pi=\hat\pi-\hat B$ is the drift-purged reading used for answer selection.
$\kappa$ is set from the erosion distribution of healthy (\sco) held-out chains.

\paragraph{Level 2 (state steering).} Let $J_t=(\nabla_h g_\phi(h_t,t))^\top$
be the $1\times d$ Jacobian row of the scalar critic. Then
$u_t=-(J_t^\top J_t+\rho I)^{-1}J_t^\top g_\phi(h_t,t)$, with $J_t^\top J_t$
and $I$ both $d\times d$, is the Gauss--Newton
step for $\min_u\lVert g_\phi(h_t+u,t)\rVert^2$ to first order; it vanishes as
$g_\phi\to0$. For a \emph{linear} critic $J_t$ does not depend on the current
hidden state, so the update follows a time-indexed fixed direction with
state-dependent magnitude---the sense in which static activation steering
\citep{li2023iti} is the degenerate case. A critic constant in $h_t$ has
$J_t=0$ and produces no steering at all. Variance-gate the intervention with the
critic's uncertainty head; fall back to Level 1, which does not modify hidden states.

\paragraph{Implementation choices for language-model training (Alg.~\ref{alg:train}).} The
$\theta$-step differentiates the penalty through generated text. Sampled tokens
are discrete, so the pathwise gradient is unavailable as written and a concrete
estimator must be chosen: policy-gradient/REINFORCE on the sampled trajectory, a
teacher-forced fixed-token surrogate, a straight-through relaxation, or
differentiable latent reasoning. Our synthetic study differentiates a continuous
belief directly and so does not settle this; the reference implementation we
release adopts the teacher-forced surrogate (sample a rollout without gradient,
re-forward it with gradient, and propagate only through the readout's dependence
on the hidden states). We therefore state Alg.~\ref{alg:train} as conceptual for
autoregressive models.

\paragraph{A frozen head is not a frozen ruler.} Freezing the readout's
\emph{parameters} prevents direct updates to the readout weights, but
updating $\theta$ changes the representation $h_t$ that feeds the readout, so
its calibration and semantics can drift even with fixed weights. Any real
fine-tuning run should therefore monitor readout calibration on held-out
problems during training, or freeze the tapped representation below the
intervention point; we do not consider this solved.

\paragraph{Feedback trap.} Never fit the critic on corrected trajectories: the
controller removes the signal the critic estimates. Audit rollouts must stay open-loop.

\begin{algorithm}[h]
\caption{Drift-penalized training (two-timescale; conceptual---see
App.~\ref{app:deploy} for the gradient estimator a language model requires)}
\label{alg:train}
\begin{algorithmic}[1]
\STATE train readout $r$ on base-model rollouts; \textbf{freeze} $r$; $\pi_t=r(h_t)$
\WHILE{not converged}
  \STATE roll out uncorrected chains; $\Delta^{(i)}_t=\pi^{(i)}_{t+1}-\pi^{(i)}_t$
  \FOR{$k=1$ to $K$}
    \STATE $\phi\mathrel{+}=\alpha_\phi\nabla_\phi\frac1n\sum_{i,t}
      \big(2g_\phi(h^{(i)}_t,t)\Delta^{(i)}_t-g_\phi(h^{(i)}_t,t)^2\big)$
  \ENDFOR
  \STATE $\theta\mathrel{-}=\alpha_\theta\nabla_\theta(\widehat\Lt+\lambda\widehat V)$
    \hfill($\phi,r$ fixed)
\ENDWHILE
\STATE \textbf{return} $\theta$; retain $g_\phi$ as the inference-time drift sensor
\end{algorithmic}
\end{algorithm}

\begin{algorithm}[h]
\caption{Drift audit / critic fit (frozen model $\theta$)}
\label{alg:audit}
\begin{algorithmic}[1]
\STATE roll out uncorrected chains on $\{q^{(i)}\}\sim\D$; record $h^{(i)}_t$,
  $\pi^{(i)}_t=r(h^{(i)}_t)$
\STATE fit $g_\phi=\arg\min_g\sum_{i,t}\big(\Delta^{(i)}_t-g(h^{(i)}_t,t)\big)^2$
  \hfill(equivalent to \eqref{eq:var})
\STATE \textbf{return} $g_\phi$; report held-out $\widehat V=\widehat\E[2g_\phi\Delta-g_\phi^2]$
  and chain-clustered uncertainty
\end{algorithmic}
\end{algorithm}

\paragraph{Training with gold anchors (future work).} Eq.~\eqref{eq:train} enforces
martingale (self-consistency) calibration; it is silent about \emph{which} verdict is
conserved. When verified $(q,a^\star)$ pairs exist, a complementary phase teacher-forces
the gold answer and supervises the belief toward the harmonic extension of the
\emph{verified} boundary with a proper scoring rule---supervising calibration, not
token-likelihood. Critic data must remain free-running (the guard above); gold labels
are load-bearing only at the boundary. Fine-tuning a full reasoning model with the
combined objective, and the accuracy-vs-budget flattening it predicts, is the
natural next experiment; we did not run it. Expected pitfalls, for those who do: (i) the frozen-ruler guard matters
most under fine-tuning, since the readout would otherwise drift with the model;
(ii) interaction with preference-tuned policies is untested---a KL-to-reference
term and the drift penalty regularize different objects and should compose, but
the two-timescale inner maximization adds instability risk; (iii) rationale style
may shift (the penalty discourages predictable belief movement, not predictable
\emph{text}); and (iv) with positive average predicted drift at these scales, the penalty's
measurable effect may be small until adverse-drift regimes (longer budgets, weaker
models, distribution shift) are targeted deliberately.

\section{Experimental Details}
\label{app:expdetails}

\paragraph{Internal pre-registration.} Gates, thresholds, operating
points, label definitions, and interpretation rules were written down and dated
\emph{before} the corresponding real-data results were computed, and are
reproduced in App.~\ref{app:expdetails}; the dated plan itself ships with the
supplement. The original gates were fixed before any real-data run, and the
secondary analyses before any Exp.~1/2 number was unblinded. The record is an
internal, dated plan rather than an entry in an external registry; we use
``pre-registered'' in that sense throughout. Its content is auditable against
the reported outcomes: Gate~1(a) failed and triggered its committed fallback,
Gate~2 was met at no operating point, and rule~(ix) suppressed a detection
claim---all reported as such.

\begin{figure}[t]
\centering
\maybefig{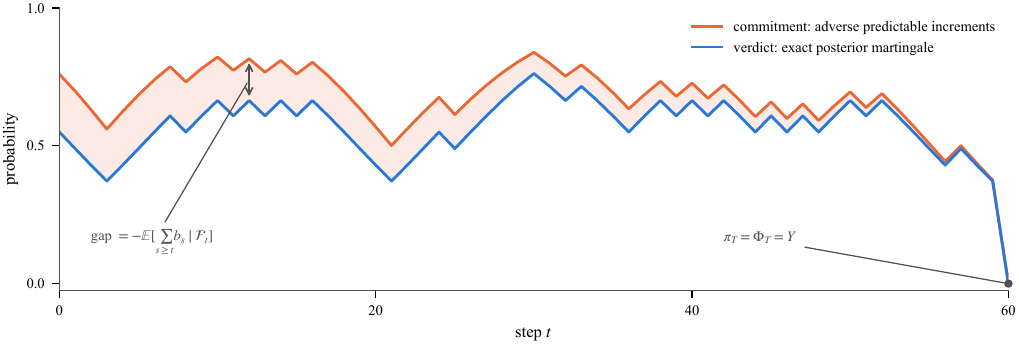}{\textwidth}{3.1cm}
\caption{The commitment--verdict gap on a simulated binary-evidence chain
with adverse predictable drift (illustrative). The verdict is the exact
posterior of a binary terminal outcome, revealed at termination. The
commitment adds a nonnegative offset whose conditional expectation decreases,
giving adverse predictable drift. The shaded gap equals
$-\E[\sum_{s\ge t}b_s\mid\F_t]$ and vanishes when both readouts equal $Y$.
It need not shrink monotonically on an individual path.}
\label{fig:schematic}
\end{figure}

\paragraph{Historical cross-protocol branching diagnostic.}
The earlier study uses the same 12 wrong-ending 1.5B chains at their first
correct checkpoint ($\le2048$; all five levels), but samples
$\brM$ immediate answers and $\brCont$ continuations with \emph{4096 additional
tokens and no terminal forcing}. Its continuation success probability
$\Psi_t$ is not the primary-policy verdict $\Phi_t$.
It gives $\hat\pi_t=\brMeanPi$, $\hat\Psi_t=\brMeanPhi$ and gap $\brMeanGap$;
$\brCapPct\%$ of $\brContN$ continuations are capped without an answer,
versus $\brCtrlCapPct\%$ of $\brCtrlContN$ in the earlier $\brCtrlN$-prefix
control arm. Low success therefore largely reflects nonresolution.
The post-hoc partition ($\hat\pi_t\ge0.75$, gap $\ge0.5$) is descriptive.
These historical results are retained below but are not substituted for
the matched follow-up in Sec.~\ref{sec:exp1}.

\begin{center}
\small
\begin{tabular}{r r c c c}
\toprule
problem & checkpoint & $\hat\pi_t$ & $\hat\Psi_t$ & gap\\
\midrule
\brPairsTable
\bottomrule
\end{tabular}
\end{center}

\begin{figure}[t]
\centering
\maybefig{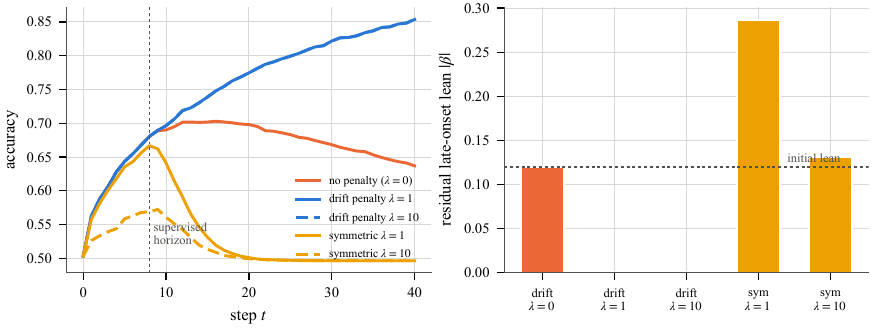}{\textwidth}{4.0cm}
\caption{Late-onset lean: the lean acts only after the supervised horizon ($t>8$;
task loss at $t=8$), so task supervision cannot see it. \textbf{Left:} accuracy vs.\
step; without a penalty the late lean erodes accuracy after the horizon; the drift
penalty restores the calibrated trajectory; the symmetric penalty flattens the gain
as well. \textbf{Right:} residual lean $|\beta|$ by objective.}
\label{fig:lateonset}
\end{figure}

\paragraph{Experiment 0: late-onset drift and the configuration sweep.}
A pre-registered late-onset test (Fig.~\ref{fig:lateonset}) makes the penalty
load-bearing: a lean beginning only after the supervised horizon survives task
supervision, is removed by the drift penalty ($\lateDriftAcc\%$ accuracy with gain intact), and defeats the
symmetric penalty, which instead saturates the belief. Across 19 synthetic
configurations, median compensation recovery is $\sweepRecMed\%$; the main failure
mode is belief saturation, where probability-scale drift becomes hard to observe.

\begin{figure}[t]
\centering
\begin{minipage}[t]{0.42\textwidth}
\centering
\maybefig{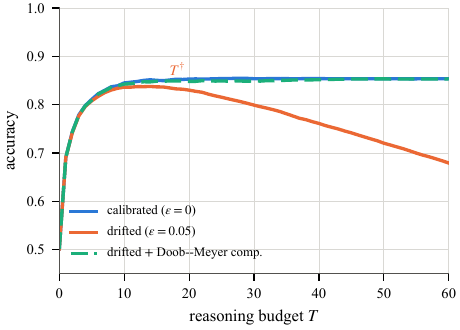}{\linewidth}{3.0cm}
\end{minipage}\hfill
\begin{minipage}[t]{0.56\textwidth}
\centering
\maybefig{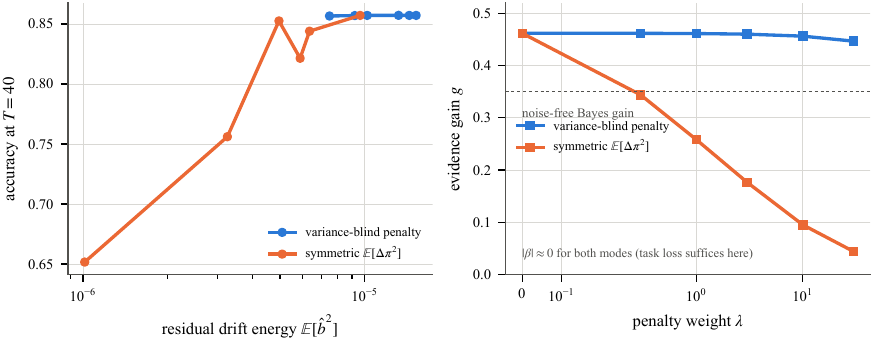}{\linewidth}{3.0cm}
\end{minipage}
\caption{\textbf{Left:} synthetic overthinking under decaying evidence and a
persistent lean; compensation recovers the calibrated plateau. \textbf{Right:}
the symmetric penalty destroys evidence gain and accuracy, while the
variance-blind penalty leaves both intact across $\lambda$.}
\label{fig:exp0panels}
\end{figure}

\paragraph{Hardware and stack.} One RTX 4070 Ti (12\,GB), Windows 11, PyTorch cu126,
Hugging Face \texttt{transformers} (SDPA attention), fp16. Primary-audit seed 0;
follow-up studies use their recorded, disjoint seed streams.

\paragraph{Rollouts.} MATH-500 test split; prompt = problem +
``\emph{Please reason step by step, and put your final answer within}
\texttt{\textbackslash boxed\{\}}.'' via the chat template (the R1-distill template
opens \texttt{<think>} automatically); temperature 0.6, top-$p$ 0.95, max 4096 new
tokens, one chain per problem, batch 8.

\paragraph{Forcing.} At checkpoints $\{512k\}_{k=1}^{8}$ (truncating mid-think):
append \texttt{\textbackslash n</think>\textbackslash n\textbackslash n**Final Answer:**
\textbackslash boxed\{} and decode greedily ($\le48$ tokens, parsed with the
permissive brace-closing rule, which accepts a continuation that never closes);
if the chain had already closed its think block, the forcing string is appended without
the closing tag. The natural termination is its own checkpoint (the chain's own boxed
answer where present). Verification: normalized string match with \texttt{math-verify} fallback;
unparseable $\Rightarrow$ incorrect.

\paragraph{States and steps.} Teacher-forced prefill; hidden states at layers 14/21/27
(of 28) at paragraph boundaries (\texttt{\textbackslash n\textbackslash n} splits,
segments $<10$ tokens merged forward, $\le96$ steps/chain) and at every checkpoint
position. Layer 21 is primary; 14/27 form the robustness panel.

\paragraph{Probes and critic.} Commitment probe: logistic regression on checkpoint
states with label = forced-answer correctness at that checkpoint; $C$ selected by inner
CV over $\{0.01,0.1,1\}$; temperature-scaled on one fifth of the probe split; 15-bin
ECE reported. Verdict probe: identical protocol, label = final-chain correctness.
Critic: LayerNorm $\to$ 2$\times$256 GELU MLP, dropout 0.1, mean and log-variance
heads, Gaussian NLL, AdamW ($10^{-3}$, wd $10^{-4}$), batch 1024, early stopping;
input $(h_t, t/4096)$; targets $\Delta\hat\pi_t$ on uncorrected rollouts of the critic
split only. Splits: 40/30/30 problem-level, stratified by (level, final correctness).

\paragraph{Labels.} \emph{\ow}: correct at some checkpoint $\le2048$ tokens (strictly
before the final answer) and wrong at the final answer; \emph{\sco}: the same but
correct finally; chains ending before the first checkpoint have no early checkpoint and
can receive neither label;
\emph{late-bloomer}: no early correct checkpoint, correct finally;
\emph{early-unresolved/wrong}: no early correct checkpoint, wrong finally.
The latter is stored as \texttt{never\_correct} in the archived code and shown
as ``never-correct'' in legacy figures. It does not imply correctness nowhere:
the implemented label does not inspect transient correctness after token 2048.

\paragraph{Conventions.} The theory's abstract ``step'' is instantiated as a
paragraph segment of the think text (typically 40--90 per chain); checkpoints are
token-indexed, and the critic receives the token position, so the two clocks are
linked through $(h_t,t)$. ``Mid-chain'' is the first paragraph step at or after
50\% of the chain's generated length (capped at the last step). Chains with fewer than four
paragraph steps are excluded from step-level analyses.

\paragraph{Probe health at the pre-registered tap.} Rule~(viii) is not triggered
at either scale, on its pre-registered AUC condition alone: the verdict probe's
final-correctness AUC is $\verdictAUCthree$ at 1.5B and $\sevBverdictAUCthree$
at 7B, neither above $0.75$. Separately---and as a post-hoc judgment, not part
of the dated rules---the 1.5B verdict probe is poorly calibrated (ECE
$\verdictECE$ against $\probeECE$ for the commitment probe), so we do not read
its predictable fraction as evidence about the tap.

\paragraph{Probe-shift and difficulty controls.} Drift energy near vs.\ far from probe-training checkpoints is comparable
($\energyNear$/$\energyFar$; 7B $\sevBenergyNear$/$\sevBenergyFar$),
which does not reveal a large checkpoint-proximity effect; on levels 1--2 (1.5B) both overthink rate and matched-position drift energy
fall ($\ncLoneTwoRate\%$ vs.\ $\ncLthreeFiveRate\%$; $\ncMatchedLoneTwo$
vs.\ $\ncMatchedLthreeFive$).

\paragraph{Mean predicted drift by group.} 1.5B: $\gdOWmean$ \ow,
$\gdSCmean$ \sco, $\gdLBmean$ late-bloomer, $\gdNCmean$ early-unresolved/wrong; 7B:
$\sevBgdOWmean$, $\sevBgdSCmean$, $\sevBgdLBmean$, $\sevBgdNCmean$.

\begin{figure}[h]
\centering
\maybefig{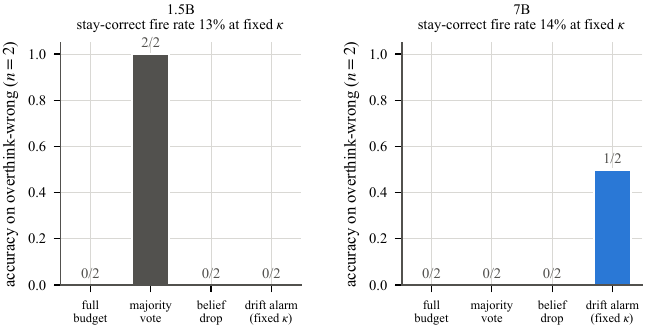}{0.85\textwidth}{3.6cm}
\caption{Per-\ow{}-chain outcomes at the pre-registered operating point (1.5B
left, 7B right; $k/n$ annotated, $n=2$ eval events per scale). Anecdotal at this
prevalence---deferred here per the interpretation rules; fire rates
$\alarmFireOW\%$ / $\sevBalarmFireOW\%$ on \ow.}
\label{fig:exp2ow}
\end{figure}

\paragraph{Per-group outcomes at the fixed operating point
(Fig.~\ref{fig:exp2ow}).} Alarm at the pre-registered $\kappa$ (90th \sco{}
percentile, critic split), eval levels 3--5:
\ow{} $\alarmAccOW\%$ ($\opAccOWstrict\%$ strict), \sco{} $\alarmAccSC\%$,
late-bloomer $\opAccLB\%$, early-unresolved/wrong $\opAccNC\%$; fire rates
$\alarmFireOW\%$ (\ow) vs.\ $\alarmFireSC\%$ (\sco); mean charged tokens
$\alarmTokens$ vs.\ $\fullTokens$ at full budget.

\paragraph{Exp 2 policies.} All replays are nonanticipating; answers only at recorded checkpoints
(the last checkpoint at or before the trigger). $\kappa\in$
$\{50,70,80,90,95\}$th percentiles of $\max_t(-\hat B_t)$ on \sco{} critic-split
chains; belief-drop thresholds $\delta\in\{0.1,\dots,0.5\}$ for both raw $\hat\pi$ and
compensated $\hat\pi-\hat B$; fixed stops at every checkpoint. Tokens are
charged by rule~(vi) below.

\begin{figure}[h]
\centering
\maybefig{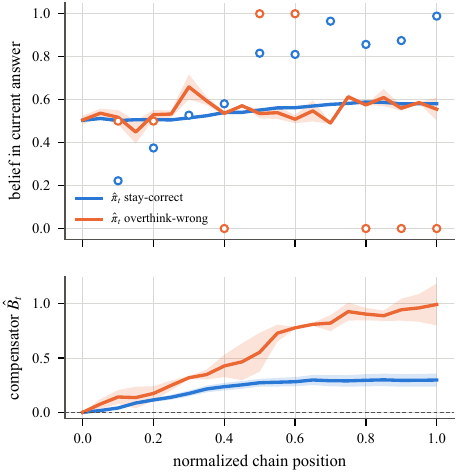}{0.5\textwidth}{4.6cm}
\caption{1.5B belief and compensator trajectories by group (eval, levels 3--5;
illustrative---two eval \ow{} chains).}
\label{fig:exp1traj}
\end{figure}

\begin{figure}[h]
\centering
\maybefig{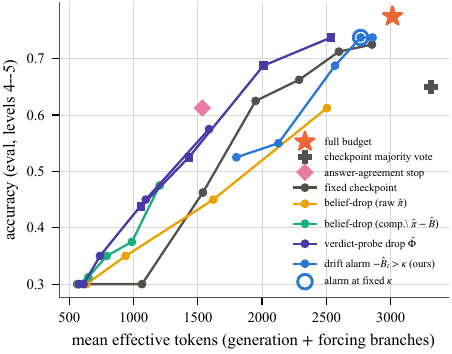}{0.9\textwidth}{4.0cm}
\caption{Stopping-policy replays at 7B (levels 4--5 eval): every family sits
below the full-budget point; the monotone cohort curve separately explains why
\emph{global truncation} has no favorable operating point.}
\label{fig:exp2}
\end{figure}

\begin{figure}[h]
\centering
\maybefig{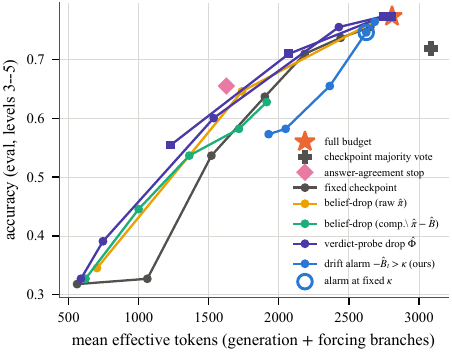}{0.9\textwidth}{4.0cm}
\caption{1.5B stopping-policy replay (levels 3--5 eval), analogous to
Fig.~\ref{fig:exp2}.}
\label{fig:exp2b}
\end{figure}

\paragraph{Escalation outcome.} Gate 1(a) failed at 1.5B
(10 \ow{} chains on levels 3--5, minimum 30); the escalation to 7B-NF4 on levels
4--5---the pre-registered fallback of Gate 1(a)---was executed with identical
protocol and returned 9 \ow{} of 262. Gate 0 passed (Sec.~\ref{sec:exp0}); Gate 2
was met at no $\kappa$ at either scale (the alarm loses accuracy at both). Both
audits are reported in full. The rule-(ix) predictor comparison is unpowered---the
evaluation split holds $\evalOW$/$\sevBevalOW$ \ow{} chains (1.5B/7B) against
$\evalSC$/$\sevBevalSC$ \sco{}, below the pre-committed minimum of three positives for reporting AUC---so no
AUC is reported and the raw paired-bootstrap differences are released with the
result files in the supplement. The 1.5B stopping-policy replay
(full $\fullAcc\%$ at $\fullTokens$ tokens; alarm $\alarmAcc\%$ at
$\alarmTokens$; vote $\voteAcc\%$; agreement $\agreeAcc\%$ at $\agreeTokens$)
mirrors the 7B replay of Sec.~\ref{sec:exp2} (Pareto fronts:
Fig.~\ref{fig:exp2} at 7B, Fig.~\ref{fig:exp2b} at 1.5B); the 1.5B
commitment-probe and compensator trajectories by group are in Fig.~\ref{fig:exp1traj}.

\paragraph{Gates (pre-committed).} Gate 0: sensor slope in $[0.9,1.1]$, $R^2>0.8$, no
false drift at $\varepsilon=0$, detection floor reported. Gate 1: (a) $\ge30$
\ow{} chains on levels 3--5, with the committed fallback ``if not, escalate to
7B-4bit on levels 4--5 with identical protocol before concluding absence''; (b)
AUC($-\hat B_{\mathrm{mid}}$) $\ge0.65$ and within $0.02$ of the PCA-32 baseline
or better; (c) onset localization consistent with the flip. Gate 2: at some
$\kappa$, recovery of $\ge50\%$ of the overthinking loss on \ow{} chains; \sco{}
accuracy change $\ge-1$\,pt with fire rate $\le10\%$; tokens saved $\ge15\%$ at
matched-or-better overall accuracy.

\paragraph{Secondary pre-registered analyses (fixed before any Exp.~1/2
number was unblinded).} The original gates above are unchanged. The following
analyses, definitions, and decision rules were fixed while the rollouts were running
and before any real-data result was computed.
\emph{(i) Operating point.} The reported alarm operating point is fixed in advance as
the 90th percentile of $\max_t(-\hat B_t)$ on \sco{} critic-split chains; other
percentiles appear only in the Pareto sweep.
\emph{(ii) Strict labels.} Because a single forced answer is a noisy draw, all
headline quantities are additionally reported under a strict label: \ow-strict
requires correctness at two consecutive checkpoints $\le2048$ with the same
normalized answer string, and a wrong final answer; \sco-strict analogously. The
population and the conclusions must survive this tightening. Per-checkpoint answer
consistency (fraction of checkpoints agreeing with the chain's modal answer) is
reported by group.
\emph{(iii) Baseline gauntlet (Exp 1).} The predictor comparison additionally
includes:
a supervised MLP (critic architecture) classifying \ow{} vs.\ \sco{} from
$h_{\mathrm{mid}}$; a length-only predictor; a Signal-A-only predictor (forced-answer
pattern up to mid-chain); an entrenchment score in the style of
\citet{he2025martingale} (per-chain regression of $\Delta\hat\pi$ on $\hat\pi$);
and critic ablations (no step-index input; input $(\hat\pi_t,t)$ only). Paired
bootstrap confidence intervals are reported for
$\mathrm{AUC}(-\hat B)-\mathrm{AUC}(\text{probe slope})$ and
$\mathrm{AUC}(-\hat B)-\mathrm{AUC}(\text{supervised MLP})$.
\emph{(iv) Prediction vs.\ postdiction.} All predictor rows are also reported
restricted to \ow{} chains whose forced answer is still correct at mid-chain.
\emph{(v) Policy gauntlet (Exp 2).} Added causal-replay policies: checkpoint majority
vote over the eight forced answers; answer-agreement stopping (two consecutive
checkpoints with the same normalized answer); verdict-probe level and drop stopping;
the fixed stop at 2048 called out against Gate~2's region; and alarm vs.\ raw-drop
detection compared at matched \sco{} fire rate.
\emph{(vi) Token accounting.} Tokens charged = trigger position plus 48 tokens per
forced-answer branch executed by the policy (a fixed allocation, not measured
decoded length); fixed stops and stop-policies are charged one branch per executed
forcing, majority voting is charged all eight, and full budget is charged none,
including when its terminal answer required forcing. Critic evaluation and
branch prefill are not charged. These are archived token allocations, not
measured wall-clock time or FLOPs.
\emph{(vii) Onset null.} Onset localization is reported against a within-chain
permutation null (random step position), with grid resolution $512$ tokens noted.
\emph{(viii) Placebo normalization and failure criterion.} The placebo is assessed as
the predictable fraction $\E[\hat m^2]/\E[\Delta^2]$ against a probe-matched
shuffled control at comparable increment variance. If the verdict probe's normalized
predictable fraction exceeds $3\times$ its shuffled control while its final-correctness
AUC exceeds $0.75$, we will report this as evidence against the tapped-state
harmonicity approximation, not as probe estimation error.
\emph{(ix) Interpretation rules.} If $\mathrm{AUC}(-\hat B)$ is statistically
indistinguishable from the supervised classifier, probe-slope, or length-only baselines
(paired bootstrap), we will report that mid-chain states generically predict failure
and that the compensator's contribution is semantic (interpretation, onset
localization, the readout--verdict gap) rather than raw detection power. If the drift
alarm does not Pareto-dominate checkpoint-majority voting and verdict-probe stopping
under accounting (vi), we will report the alarm as one member of a family of effective
stopping rules rather than as superior.
\emph{(x) Resampling study.} If compute allows, $k=4$ additional chains are sampled
for $\sim$100 level-3--5 problems to measure label stability under resampling, the
per-problem accuracy-vs-budget estimand, a sibling-chain shared-drift test, and a
token-matched self-consistency point for the Pareto curve.

\paragraph{Preliminary AIME-2024 audit (post-hoc).} As a
hardest-regime probe we ran $k=4$ chains per problem on all 30 AIME-2024 problems at
1.5B with the identical protocol ($\aimeChains$ chains; final accuracy
$\aimeFinalAcc\%$; $\aimeCapRate\%$ hit the 4096-token cap). No chain is \ow{} ($\aimeOW$ of $\aimeChains$) and the capped-cohort curve
rises monotonically ($\aimeCurveStart\%\to\aimeCurveEnd\%$,
$n=\aimeCappedN$), but this configuration is overwhelmingly
competence-limited: 99 chains are never correct, and only $\aimeElig$ of
$\aimeEligN$ chains ever reach a correct early checkpoint and are therefore
\emph{eligible} for an \ow{} transition at all. On that opportunity set the
count is $\aimeOW$/$\aimeElig$, whose 95\% upper bound is
${\approx}\aimeUpper\%$. The zero is a low-powered null, not evidence that
overthinking is absent where folklore expects it; budgets beyond 4096 remain
untested.

\paragraph{Forcing-operator validation (post-hoc).} To test
whether the budget-forced answer matches a \emph{natural} halt, we truncated
$\fvPairs$ stratified (chain, checkpoint) pairs at 1.5B, closed the think block
\emph{without} the forcing string, and let the model continue greedily for up to
220 tokens. Most natural continuations ($\fvNoAns\%$) emitted no boxed answer at
all---absent the forcing string the model tends to resume reasoning rather than
commit, which is why the operator is needed in the first place. Where a natural
answer \emph{did} appear, it matched the forced answer string in $\fvAgree\%$ of
cases and agreed on correctness in $\fvCorrAgree\%$; the residual disagreement is
consistent with genuine indeterminacy of mid-chain commitment under two different
halting prompts. We therefore read $\hat\pi_t$ as commitment \emph{under the
standard budget-forcing operator} (the object our conventions define), with agreement measured only among the minority of prefixes that answer
under both protocols; this does not bound unconditional operator error.

\paragraph{Checkpoint-grid robustness (post-hoc).} Recomputing
labels on a coarser 1024-token grid (a strict subset of recorded forcings):
\ow{} counts move $\gridFineOW\to\gridCoarseOW$ at 1.5B and
$\gridFineOWsevB\to\gridCoarseOWsevB$ at 7B. Because the grids are nested and
the \ow{} label requires a correct answer at \emph{some} checkpoint, refining
the grid can add chains but never remove one: the count is non-decreasing in
density, so the ${\approx}3\%$ figure is specific to the 512-token grid---a
ceiling for coarser grids and a floor for finer ones, not a one-sided bound.

\paragraph{The $k{=}4$ quota resample is not a prevalence estimate.} We do
\emph{not} pool the $k{=}4$ quota-selected resample set (item~x) into a
prevalence estimate; the pooled estimate is the separate $k{=}8$ study below.
That set was selected by outcome-dependent quotas on the original
chain's label, so it is a census of the $\wpOWpop$ \ow{} problems and is
enriched ${\approx}\wpEnrich\times$ for the outcome being estimated; that
design supports label stability, not prevalence. Post-stratifying by the known
population strata gives $\wpRate\%$ $\wpCI$ for a fresh chain (against
$\wpUnw\%$ unweighted on the enriched sample), which merely recovers the
independent single-chain rate $10/367$ and so adds little information---the
interval is wide because the \ow{} stratum contains only $\wpOWpop$
problems.

\paragraph{Resampling study (pre-registered item x; results).} $k=4$ fresh chains
per problem on $\rsN$ level-3--5 problems (1.5B, identical protocol). Label
stability: $\rsOWany$/$\rsOWorig$ \ow{} problems reproduce the label in at least
one sibling; $\rsSCstable$/$\rsSCorig$ \sco{} problems remain \sco. The
sibling-chain shared-drift test finds no problem-level adverse-drift signal (the
original chain's mid-chain erosion does not predict sibling overthinking; Pearson
$r=\rsSharedR$), and self-consistency over the five chains ($\rsSCacc\%$ at
$\rsSCtokens$ tokens) does not beat one full-budget chain here. This quota
sample is too small to settle whether flips are chain-idiosyncratic; the larger
$k{=}8$ study below, on the complete problem set, finds a clear problem-level
propensity ($\rEightSibOW\%$ of \ow{} siblings against $\rEightSibSC\%$ of
\sco{} siblings) alongside substantial chain noise, and supersedes this reading.

\paragraph{$k{=}8$ resampling (pre-registered).} Eight fresh chains per
problem, from eight disjoint seed streams, on all 367 level-3--5 problems at
1.5B, under the audit protocol otherwise unchanged ($T{=}0.6$, top-$p$ $0.95$,
4096-token cap, 512-token forcing grid, identical labels): $\rEightN$ chains, of
which $\rEightOW$ \ow{}, $\rEightSC$ \sco{}, $\rEightLB$ late-bloomer and
$\rEightNC$ early-unresolved/wrong. Unlike item~(x) the problem set is complete rather
than quota-selected, so pooling is legitimate: the rate $\rEightRate\%$
(problem-clustered 95\% CI $\rEightCI$) is the repeated-sample prevalence in
Sec.~\ref{sec:exp1}, which also reports the single-chain counts.
As the dated plan specifies, it supersedes the
single-chain $10/367$ as the headline. Sibling propensity and the
non-monotonicity screen are documented in this appendix.

\begin{figure}[t]
\centering
\maybefig{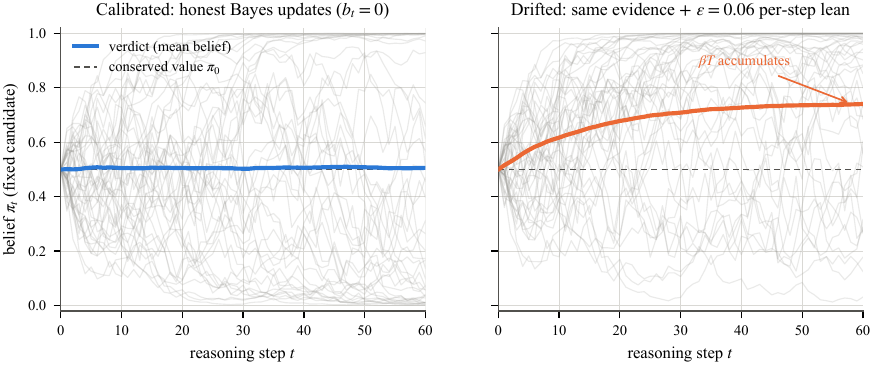}{0.78\textwidth}{2.6cm}
\caption{Bayesian evidence accumulation (belief $=$ posterior on a fixed
candidate). \textbf{Left:} honest updates ($b_t=0$): chains diffuse to $0/1$,
the mean belief ($=$ verdict) is conserved (max deviation $\eZeroMeanDev$).
\textbf{Right:} the same evidence with a constant lean: the mean bends toward a
confidently wrong value, ballistically.}
\label{fig:verdict}
\end{figure}

\paragraph{Exp 0 details.} Evidence separation $d=0.35$, $\sigma=1$, $T=60$
($\varepsilon=0.06$ in Fig.~\ref{fig:verdict}); the drift arm adds $\varepsilon$ to
each log-odds increment (probability-scale drift $\approx\varepsilon\pi(1-\pi)$;
slope/$R^2$ are measured against the exact $b(h,t)$ by quadrature, not raw
$\varepsilon$). Overthinking variant: initial evidence scale $1.0$ decaying as
$e^{-t/8}$ with the per-step Bayes gain, lean persistent. Objective panel: $T=40$,
learnable (gain, lean), fixed representational noise $\tau=0.12$ (hence the learned
gain $\approx0.46$ above the noise-free Bayes gain $0.35$), scarce-label task loss (96
labeled problems), penalty on unlabeled rollouts, $\lambda\in\{0,0.3,1,3,10,30\}$,
critic 5 inner steps per reasoner step.

\subsection{Post-hoc robustness analyses (not pre-registered)}
\label{app:audit2}
The analyses in this subsection were run \emph{after} unblinding, on the frozen
rollout and state caches. They are sensitivity checks on the pre-registered
conclusions, not new pre-registered claims.
\paragraph{Survivorship decomposition: empirical terms (Prop.~\ref{prop:decomp}).}
Conditioning follows the published curve's convention ($R(T)$: checkpoint $T$
recorded, $=\{\tau\ge T\}$ on the grid, matching Prop.~\ref{prop:decomp};
capped chains included at the cap); the
reproduced aggregate curve matches the published one to machine precision and the
identity is exact per step. Region totals over $2048\to4096$ (chain-level
bootstrap, 2000 resamples): 1.5B marginal change $\decompTotal$ $=$ within
$\decompWithin$ $\decompWithinCI$ $+$ composition $\decompComp$
$\decompCompCI$; 7B $\sevBdecompTotal$ $=$ $\sevBdecompWithin$
$\sevBdecompWithinCI$ $+$ $\sevBdecompComp$ $\sevBdecompCompCI$ points. The
within-cohort term is significantly \emph{positive} at both scales---chains
still reasoning keep improving---so survivorship more than explains the entire
marginal decline.

\begin{center}
\small
\begin{tabular}{r rrrr c rrrr}
\toprule
& \multicolumn{4}{c}{1.5B} & & \multicolumn{4}{c}{7B}\\
\cmidrule{2-5}\cmidrule{7-10}
$T$ & $n$ & $M(T)\%$ & within & comp & & $n$ & $M(T)\%$ & within & comp\\
\midrule
\decompTable
\bottomrule
\end{tabular}
\end{center}

\paragraph{Label-definition robustness.} Relaxing the strict pre-registered
\ow{} definition: dropping the $\le2048$-token requirement gives
$\lenAny$/$367$ ($\lenAnyPct\%$) at 1.5B and $\sevBlenAny$/$262$
($\sevBlenAnyPct\%$) at 7B---at 7B identical to the base count: every
early-correct-then-wrong chain was already correct by token 2048. Requiring two
consecutive correct checkpoints gives $\lenSus$ ($\lenSusPct\%$) and
$\sevBlenSus$ ($\sevBlenSusPct\%$). Counting chains with \emph{any}
correct$\to$wrong forced-answer flip anywhere gives $\lenFlip$
($\lenFlipPct\%$) and $\sevBlenFlip$ ($\sevBlenFlipPct\%$)---but most
flipped chains recover by the natural end. Prevalence stays within
${\sim}1$--$4\%$ under every final-outcome definition.

\paragraph{Cumulative-history-summary ablation.}
This follow-up was specified before its own results, but after the original
audit. Fixed-window taps can forget information: a pair tap at $t+1$ no
longer explicitly includes $h_{t-1}$. Moreover,
\texttt{multilayer\_tap.py} enriches only the critic in its pair variant,
leaving both probes on single states. We therefore refit both probes and
both critics with $z_t=[h_t,\mathrm{mean}(h_0,\ldots,h_t)]$ at layer 21.
The splits, seed, calibration fraction, hyperparameters, and shuffled-label
diagnostic are unchanged. A single-state control reproduces
$\ntSingleCommit$/$\ntSingleVerdict$ and
$\sevBntSingleCommit$/$\sevBntSingleVerdict$.

\emph{This construction is history-dependent but not necessarily nested.}
For example, histories $(h_0,h_1,h_2)=(0,1,0)$ and $(1,0,0)$ give different
$z_1$ but identical $z_2=(0,1/3)$. Hence the construction does not guarantee
$\sigma(z_t)\subseteq\sigma(z_{t+1})$.
The filtration $\sigma(z_0,\ldots,z_t)$ is nested; conditioning on $z_t$
alone is a different operation. A compressed recurrent state likewise
requires a sufficiency argument before inheriting the verdict martingale.

The verdict diagnostic decreases from $\ntSingleVerdict$ to
$\ntNestedVerdict$ at 1.5B and from $\sevBntSingleVerdict$ to
$\sevBntNestedVerdict$ at 7B, while the commitment diagnostic also falls
($\ntSingleCommit$ to $\ntNestedCommit$; $\sevBntSingleCommit$ to
$\sevBntNestedCommit$). The planned positive commitment-minus-verdict gap
is not obtained ($\ntNestedGap$/$\sevBntNestedGap$). These results show only
that this augmentation does not resolve the control discrepancy; they
cannot distinguish observation loss from estimation error. Nor does the
tower property require this ordering for fitted, variance-normalized
diagnostics.

There are additional confounds. The doubled input dimension raises the
7B commitment shuffled-label diagnostic from $\sevBntSingleCommitNull$ to
$\sevBntNestedCommitNull$ (42\% of the measured value), limiting comparisons
across feature sets. The implementation computes running means over
checkpoint states for probe fitting but over paragraph-step states for the
increment series, so the history-summary sampling rules differ between
training and application. A conclusive history control would require
consistent observation sequences, an explicitly nested conditioning
history, and separate assessment of estimation error. The archived
\texttt{nested\_tap} filenames and numerical macros retain their original
names for traceability, not as a claim of nested information.

\paragraph{Critic capacity.} Replacing the MLP critic with a single linear
layer (identical protocol and splits): at 1.5B the linear critic retains genuine
held-out skill on increment prediction ($R^2=\linRtwo$ vs.\ the MLP's
$\mlpRtwo$), so the predictable-drift signal is not an expressivity artifact of
the MLP. The energy-ratio metric is not meaningful for the linear critic (its
shuffled-label null inflates), and at 7B the $3585$-dimensional linear map
overfits its ${\sim}4.3$k training rows ($R^2=\sevBlinRtwo$; high estimator
variance)---the 7B half of this ablation is uninformative, and the 1.5B half
carries the conclusion.

\paragraph{State-richness series under the variational score.} Commitment
and verdict variance-normalized scores as the tap is enriched (identical protocol
and splits; \emph{single} $=$ pre-registered layer, $+$\emph{history} $=$
one-step hidden-state history, \emph{3 layers} $=$ L14/21/27 concatenation),
reported as $\widehat V/\widehat{\mathrm{Var}}(\Delta)$ rather than
$\widehat{\E}[g^2]/\widehat{\mathrm{Var}}(\Delta)$, with the score difference from a \emph{paired}
chain-level bootstrap (one shared resample index for both readouts, 2000
resamples):

\begin{center}
\footnotesize\setlength{\tabcolsep}{2.5pt}
\begin{tabular}{l cc cc}
\toprule
& \multicolumn{2}{c}{1.5B} & \multicolumn{2}{c}{7B}\\
\cmidrule(lr){2-3}\cmidrule(lr){4-5}
tap & c/v & gap [95\% CI] & c/v & gap [95\% CI]\\
\midrule
\tapGapTable
\bottomrule
\end{tabular}
\end{center}

A positive commitment-minus-verdict gap is what an idealized, accurately
estimated verdict control would suggest; the tower property alone does not
imply this ordering for fitted, variance-normalized scores. The gap is
significant only at the 1.5B single-layer tap; elsewhere the gap straddles zero or, at the 7B
one-step-history tap, has a CI excluding zero in the reversed direction (one of
eight post-hoc cells, uncorrected). The 7B three-layer-plus-history row is
degenerate: its critic diverges at $21{,}505$ input dimensions and yields a
strongly negative variational value. A labeled post-hoc rerun at a smaller
learning rate gives gap $+0.015$ [$-0.020,+0.052$]. Bootstrap intervals are
conditional on the fitted models; they do not include probe or critic refit
uncertainty, and enrichment need not improve a finite-sample estimate.

\paragraph{Randomized refits are diagnostics, not calibrated hypothesis tests.}
The archived analysis used 200 refits per scheme at the primary taps and 50
at richer taps: a global training-step shuffle and a circular shift of the
\emph{concatenated} training target vector. The latter is not a within-chain
shift. Refits used at most 80 epochs, versus 200 for the observed critic.
Neither scheme establishes exchangeability for the heterogeneous,
nonstationary chains, so its exceedance ranks are not calibrated permutation
$p$-values; we make no significance claim from them. The released implementation
also provides a separately versioned within-chain shift with matched fit
settings; that correction has not been run for the results reported here.
Even within-chain shifts require additional stationarity or exchangeability
assumptions for inferential use. The positive held-out estimates and their
chain-bootstrap intervals, rather than these ranks, support the reported
probe-increment predictability.

The older $\E[g^2]$ statistic has a positive estimation-noise floor: its ratio
to the shuffled control varies across refits. The population variational
identity avoids that particular inflation, but does not make the empirical
estimate a guaranteed bound or require every randomized-refit value to be
negative. All archived numeric diagnostics are retained in the result files.

\paragraph{Base-family audits: R1-Distill-Llama 8B and Phi-4-mini-reasoning
(endpoints fixed in the dated plan before either run).} The Qwen3 audit
varies training recipe but not base lineage, so we repeated the audit on two
reasoning models with non-Qwen bases: DeepSeek-R1-Distill-Llama-8B
\citep{deepseek2025r1}, on a Llama-3.1 base, in 4-bit NF4 as in the 7B
escalation; and the 3.8B Phi-4-mini-reasoning \citep{phi4mini}, on a Phi-4
base, in fp16. Both models emit an explicit thinking block, which the forcing operator
requires. The pipeline is otherwise unmodified---one chain per problem on the
same $\llamaN$ level-3--5 MATH-500 problems, temperature $0.6$, $4096$-token
cap, the same $512$-token forcing grid and labels. Neither audit is floored or
ceilinged (final accuracy $\llamaAcc\%$ and $\phiAcc\%$). Labels: $\llamaOW$
\ow{} ($\llamaOWstrict$ strict), $\llamaSC$ \sco, $\llamaLate$ late-bloomer,
$\llamaNever$ early-unresolved/wrong for Llama; $\phiOW$ ($\phiOWstrict$),
$\phiSC$, $\phiLate$, $\phiNever$ for Phi. Both headline findings reproduce
outside the Qwen lineage. Prevalence is $\llamaPct\%$ $\llamaCI$ and
$\phiPct\%$ $\phiCI$, just below the ${\approx}3\%$ of the two primary
Qwen-based R1-distills. The
marginal decline over $2048\to4096$ ($\llamaTotal$ and $\phiTotal$ points)
again decomposes into a \emph{positive} within-cohort term ($\llamaWithin$
$\llamaWithinCI$; $\phiWithin$ $\phiWithinCI$) against composition
($\llamaComp$ $\llamaCompCI$; $\phiComp$ $\phiCompCI$). Deviations are recorded
in \texttt{results\_llama8b/} and \texttt{results\_phi4mini/}: the generation
batch was reduced from the pipeline default because a larger batch exceeded the
12\,GB card---on this platform the driver spills to host memory rather than
raising an error, so the reduction was triggered by a throughput floor---and
each run was interrupted and resumed through the pipeline's own resume logic
(on \mbox{(problem, checkpoint)} for forcing, which decodes greedily and is
therefore deterministic; rollout resumption re-seeds as the Qwen3 run did).
Every record was generated exactly once. These runs measure the two
pre-registered endpoints only: they extract no hidden states, so no probe,
critic or stopping replay is fitted for them and they do not appear in
Table~\ref{tab:stopping}. They are two additional families, not a scaling
study.

\paragraph{Third-model audit: Qwen3-1.7B (endpoints fixed in the dated plan
before the run; post-hoc relative to the original audit).} Qwen3
shares a base family with our R1-distills but is independently trained and not
R1-distilled, so this varies training recipe and distillation status, not base
lineage; we do not read it as evidence of cross-family generality. The full
pipeline---one chain per problem on the same $\qwenN$ level-3--5 MATH-500
problems, temperature $0.6$, $4096$-token cap, the same $512$-token forcing grid
and labels---applied to Qwen3-1.7B \citep{qwen3}. The audit is informative rather than floored or ceilinged (final accuracy
$\qwenAcc\%$, $63\%$ at 2048). Labels:
$\qwenOW$ \ow{} ($\qwenOWstrict$ strict), $\qwenSC$ \sco, $\qwenLate$
late-bloomer, $\qwenNever$ early-unresolved/wrong. Both headline findings reproduce:
prevalence $\qwenPct\%$ $\qwenCI$ brackets the ${\approx}3\%$ of the two
R1-distills, and the marginal decline over $2048\to4096$ ($\qwenTotal$ points)
again decomposes into a \emph{positive} within-cohort term ($\qwenWithin$
$\qwenWithinCI$) against composition ($\qwenComp$ $\qwenCompCI$). Deviations,
recorded in \texttt{results\_qwen3/qwen3\_audit.json}: generation batch 6
rather than 8 (VRAM headroom; throughput only), and the rollout was interrupted
by external process kills three times and resumed via the pipeline's resume
logic, so the sampling stream is not bit-identical to one uninterrupted seed-0
run---all $\qwenN$ chains are complete, deduplicated, and generated once.

\paragraph{A forcing-free view: length--accuracy curves are
censoring-dominated (post-hoc).} The decomposition above reads forced answers.
As a check that the forcing operator is not manufacturing the result, we repeat
the exercise with no forcing in the analysis: bin each chain by its natural
termination length $L$ and score it by its \emph{own} unforced final answer.
The familiar declining curve appears, steeply ($\nlAggSlope$ points per
512-token bin at 1.5B, $\sevBnlAggSlope$ at 7B). Two findings follow, and the
first does \emph{not} match the mechanism our forced decomposition isolates.
(i)~Stratifying by MATH level removes little of the decline---pooled
within-level slopes are $\nlWithinSlope$ / $\sevBnlWithinSlope$, so difficulty
composition accounts for only ${\approx}\nlCompPct\%$ /
${\approx}\sevBnlCompPct\%$ of it. (ii)~It is almost entirely
\emph{censoring}: restricting to chains that actually terminated ($L<4096$)
collapses the slope to $\nlNatSlope$ $\nlNatSlopeCI$ and $\sevBnlNatSlope$
$\sevBnlNatSlopeCI$. The long-length bins are populated by chains that hit the
cap without ever committing to an answer ($\nlCensPct\%$ / $\sevBnlCensPct\%$
of chains). Correspondingly, the surviving pool degrades by $\nlSurvDelta$ /
$\sevBnlSurvDelta$ points over $2048\to4096$ under this natural-answer
scoring---a same-direction, larger effect than the forced $M(T)$ moves, though
a different estimand (it carries no ``what would it answer mid-chain''
content). \emph{Caveat:} for capped chains the ``final'' answer must itself be
elicited by forcing, so this analysis is strictly forcing-free only on the
terminated subset---which is exactly where the effect is smallest. Reading:
aggregate accuracy-vs-length plots are dominated by \emph{which chains occupy
which bin}---here through censoring, in Fig.~\ref{fig:budget} through
termination-induced composition. Both are selection; neither is within-chain
erosion.
The forced decomposition remains the stronger design, since it conditions on
chain identity and needs no difficulty proxy at all.

\paragraph{Decoding-matched forcing (pre-registered).} Chains are
sampled ($T{=}0.6$, top-$p$ $0.95$) but forced answers are decoded greedily, so
the readout is taken under a different rule than the process it describes. We
re-forced every checkpoint of the $\dmN$ chains above under the chain's own
decoding rule, $\dmK$ draws each, and compared paired on identical
(chain, checkpoint) pairs. The difference is negligible: greedy exceeds sampled
forced-answer correctness by $\dmDelta$ points, 95\% problem-clustered CI
$\dmCI$ over $\dmProblems$ problems, with per-checkpoint deltas between
$-0.3$ and $+2.4$ points. \ow{} counts are $\dmOWgreedy$ under greedy and
$\dmOWsampled$ across the three sampled draws. We recorded in advance the
prediction that greedy forcing would \emph{inflate} the \ow{} count, making
the reported prevalence an upper bound in that respect; \textbf{that prediction
is not borne out}---the sampled counts are if anything marginally higher, and
the spread across draws exceeds the greedy-sampled difference. The defensible
reading is that the operator's decoding rule does not materially affect the
measurement, not that either rule is conservative.

\paragraph{Dense-grid test (pre-registered).} We recorded in advance
the prediction that a \emph{finer} forcing grid weakly increases the \ow{}
count, because each extra checkpoint is another chance to catch a transient
correct answer---the same mechanism that made the coarse-grid recount lower the
count. Re-forcing the $T{=}0.6$ chains of the previous paragraph on a
256-token grid (a strict superset of the 512 and 1024 grids) confirms it:
$\gCoarseOW$ \ow{} chains at 1024 tokens, $\gMidOW$ at 512, and $\gDenseOW$ of
$\gDenseN$ at 256. The crosstab isolates the mechanism exactly: all $\gDenseKept$
chains labeled \ow{} on the 512 grid keep the label, and the increase comes
entirely from $\gDenseGained$ chains that the coarser grid had recorded as
\emph{never-correct}---the finer grid finds a brief correct answer between
512-token checkpoints that is subsequently lost. Prevalence measured this way is
therefore non-decreasing in checkpoint density (unchanged from 1024 to 512, up
at 256), so the 512-grid figure is a floor relative to finer nested grids and a
ceiling relative to coarser ones---grid-specific rather than conservative.

\paragraph{Decoding-temperature ablation (pre-registered).} On a
stratified 120-problem subset with $k{=}2$ chains per problem, we repeat the
audit at the pipeline default ($T{=}0.6$) and at $T{=}0.9$, everything else
identical. Prevalence does not rise with temperature: $\tmpAltOW$ of
$\tmpAltN$ chains are \ow{} at $0.9$ ($\tmpAltRate\%$, $\tmpAltCI$)
against $\tmpDefOW$ of $\tmpDefN$ at $0.6$ ($\tmpDefRate\%$,
$\tmpDefCI$)---if anything lower, though the intervals overlap and this subset
is small. The structural finding reproduces at both settings: the marginal
curve peaks near 1.5--2k tokens and declines to the cap
($\tmpDefPeak\%\to\tmpDefEnd\%$ at $0.6$;
$\tmpAltPeak\%\to\tmpAltEnd\%$ at $0.9$) while the final-step within-cohort
delta stays non-negative---$\tmpDefWithin$ points $\tmpDefWithinCI$ at $0.6$
and $\tmpAltWithin$ $\tmpAltWithinCI$ at $0.9$, the latter significantly
positive. Per the pre-registration this is a sensitivity report, not a gate.

\paragraph{Probe time-input ablation (post-hoc).} The probes receive the
tapped state only, while the critic additionally receives normalized token
position, which raises the question whether the verdict probe's residual
predictable movement is an omitted-time artifact. The tested time-input
augmentations do not explain it away. Normalized
position is linearly decodable from the tap itself ($R^2=0.75$--$0.92$
out-of-sample), so the probes were never time-blind; refitting both probes with
position added (and with a scaled polynomial in position, to rule out the
feature being crushed by the shared penalty) moves the verdict fraction by
$+0.011$/$+0.004$ at 1.5B and $\le0.001$ at 7B, all smaller than the
$0.014$--$0.037$ standard deviation of the critic refit itself; and a critic
given \emph{time alone} reaches a fraction of only $0.001$--$0.006$. Probe
AUC/ECE reproduce the published values exactly in the no-time arm.

\paragraph{Head-to-head with published stopping rules (post-hoc).} We replay
named methods over the recorded chains under a single effective-token
accounting: budget forcing / fixed truncation \citep{muennighoff2025s1},
self-consistency \citep{wang2023selfconsistency}, Adaptive-Consistency
\citep{aggarwal2023adaptive}, ESC \citep{li2024esc}, First-Finish Search
\citep{agarwal2025ffs}, answer-consistency \citep{liu2025answerconvergence},
Dynasor-CoT \citep{fu2025certaindex}, ES-CoT \citep{mao2025escot}, our alarm,
and an oracle bound. At 1.5B each method is labeled \emph{pure replay}
($\blNpure$/$\blNmethods$), \emph{approximate replay} ($\blNapprox$, with the
deviation recorded per row), or \emph{not implementable} ($\blNnotimpl$); the 7B
replay has $\sevBblNmethods$ rows, $\sevBblNimpl$ of them implementable.
Confidence- and entropy-gated exits (e.g.\ \citealp{kadavath2022know}) fall in
the last class: we cached hidden states and answer strings but not token
logits, so replaying them would require new generation, and we do not
approximate them. Two step-level rules (ES-CoT, answer-consistency) never fire
at their published hyperparameters because our $512$-token forcing grid is
${\approx}10\times$ coarser than a reasoning step; this is a limitation of our
cache, not evidence against those methods, and grid-rescaled variants are
reported as \emph{our} rescaling. Outcome: $\blNbeatCheaper$ methods beat the
full budget at lower cost; the accuracy gains that survive bootstrap are all
multi-sample and cost $3$--$5\times$ the tokens. Against a token-matched fixed
truncation our alarm is within $\blAlarmVsFixed$ points---i.e.\ at equal
compute it is not better than truncating, which we regard as the sharpest
negative result about our own controller.

\paragraph{Self-consistency inherits the survivorship structure.} Majority-vote
and best-of-$N$ budget curves over the resampled MATH chains ($k{=}4$ plus the
original) and AIME chains ($k{=}4$), with endpoints pre-registered
in the supplement: the vote curve never declines within
cohorts beyond noise anywhere on the grid, and over $2048\to4096$ its pooled
within-cohort term is $\scVoteWithin$ points $\scVoteWithinCI$ against
composition $\scVoteComp$ $\scVoteCompCI$ (problem-level bootstrap, 2000
resamples); best-of-$N$ shows the same structure ($\scBonWithin$ within,
$\scBonComp$ composition). Thus changing-cohort selection also affects
vote curves; this does not establish shared adverse drift or exclude benefits
from voting at other operating points. Caveats: 80$+$30 problems give
wide intervals, and the AIME split is nearly degenerate (one problem leaves the
survivor set by the cap).

\paragraph{Alarm-threshold sensitivity.} Sweeping the alarm's calibration
percentile over $\{50,60,70,75,80,85,90,95,99\}$ with everything else fixed:
the accuracy-vs-tokens trade-off is monotone at both scales and no threshold
improves accuracy over the full budget. The sweep's nominal best
($\thrBestPct$th percentile: $\thrBestDelta$ points at 1.5B,
$\sevBthrBestDelta$ at 7B) is degenerate---it ties the full budget by (almost)
never firing---so the pre-registered 90th-percentile operating point is not
beaten by any non-trivial alternative, and the absence of an observed accuracy-improving alarm operating point
persists across this threshold sweep.

\begin{figure}[h]
\centering
\maybefig{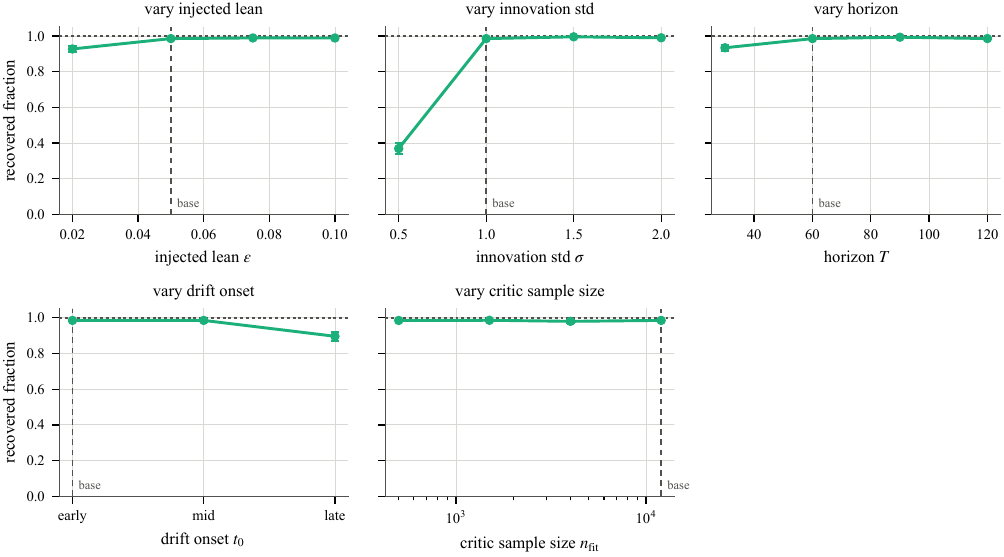}{\textwidth}{4.4cm}
\caption{Recovered fraction of injected accuracy loss under Doob
compensation, varying one factor at a time around the Exp~0 base configuration
(marked); error bars are s.e.\ over $\sweepSeeds$ seeds. Recovery is high
across all factors except very low innovation noise, where beliefs saturate.}
\label{fig:exp0sweep}
\end{figure}

\paragraph{Synthetic mechanism validation across regimes.}
One-factor-at-a-time around the Exp~0 base configuration (drift strength,
innovation noise, horizon, drift onset, critic training-sample size;
$\sweepConfigs$ grid points $\times$ $\sweepSeeds$ seeds): Doob
compensation recovers a median $\sweepRecMed\%$ of injected accuracy loss
(range $\sweepRecMin$--$\sweepRecMax\%$), robust down to $500$
critic-training problems (Fig.~\ref{fig:exp0sweep}). The one failure mode is
low innovation noise ($\sigma=0.5$, recovery $\sweepRecMin\%$): beliefs
saturate near $0/1$, where probability-scale drift vanishes and the compensator
cannot flip wrong-side saturated paths---the same saturation pathology that
defeats symmetric magnitude penalties (Sec.~\ref{sec:exp0}), here appearing as
a limit of compensation itself.

\section{Observation Histories, Loss Hazards, and Stopping Gain}
\label{app:observation}

These consequences of conditional expectation clarify what the measured
conservation property can mean; they are not proposed as new projection or
optional-stopping theorems.

\begin{proposition}[An observation-calibrated verdict can drift]
\label{prop:compressed}
Let $Y\in[0,1]$, $\mathcal G_t\subseteq\F_t$, and
$v_t=\E[Y\mid\mathcal G_t]$. The observation fields $\mathcal G_t$ need not
be nested. Write $\mathsf P_t X=\E[X\mid\mathcal G_t]$ and
$e_t=v_t-\Phi_t$, where $\Phi_t=\E[Y\mid\F_t]$. Then
\[
 m_t^{\mathcal G}:=\E[v_{t+1}-v_t\mid\mathcal G_t]
 = (\mathsf P_t\mathsf P_{t+1}-\mathsf P_t)Y
 =\E[e_{t+1}\mid\mathcal G_t],\qquad
 \|m_t^{\mathcal G}\|_2\le\|e_{t+1}\|_2.
\]
If $\mathcal G_t\subseteq\mathcal G_{t+1}$, the conditional drift is zero.
Without nesting, exact observation-level calibration alone does not imply
this conclusion.
\end{proposition}
\begin{proof}
The first identity substitutes the two conditional expectations and uses
$\mathsf P_t^2=\mathsf P_t$. For the second,
$\E[\Phi_{t+1}\mid\mathcal G_t]=\E[Y\mid\mathcal G_t]=v_t$ by
$\mathcal G_t\subseteq\F_t\subseteq\F_{t+1}$. Conditional expectation
contracts $L^2$ norms. Nesting gives the final claim by the tower property.
\end{proof}

In particular, $\mathcal H_t=\sigma(z_0,\ldots,z_t)$ is a filtration and
$\E[Y\mid\mathcal H_t]$ is a bounded martingale, even if
$\mathcal H_t\subsetneq\F_t$. Retained observation history is therefore
another sufficient route to an exact verdict null; full-state sufficiency
is not necessary. This does not make a learned history probe exact, nor does
it identify why the real-model control is nonzero.

\paragraph{Exact four-path control.}
Draw $Y\sim\mathrm{Bernoulli}(1/2)$ and a signal $S$ with
$\Pr(S=Y\mid Y)=4/5$. At $t=0$ the observation is $S$; at $t=1$ it is
constant; at $t=2$ it is $Y$. The full history retains $S$ throughout.
Conditional on $S=1$, the full verdict is $(4/5,4/5,Y)$ and the compressed
verdict is $(4/5,1/2,Y)$; for $S=0$ they are $(1/5,1/5,Y)$ and
$(1/5,1/2,Y)$. Every compressed value is exactly $\E[Y\mid z_t,t]$.
Its first conditional increment is $-3/10$ or $+3/10$, while its second
has zero conditional mean given the current constant observation. Averaged
over the two steps, predictable squared energy is $9/200$, compared with
zero for the full verdict. Retaining $S$ at $t=1$ restores the null.
All four path probabilities ($2/5,1/10,1/10,2/5$) are enumerated with exact
rational arithmetic. This is a possibility result about measurement, not
evidence that forgetting causes the LLM verdict residual.

\begin{proposition}[Binary commitments and loss hazards]
\label{prop:hazard}
Let $A_t\in\{0,1\}$ be the exact correctness of the fixed greedy stopping
operator, adapted to $\F_t$. Set
$p_{10,t}=\Pr(A_{t+1}=0\mid\F_t)$ and
$p_{01,t}=\Pr(A_{t+1}=1\mid\F_t)$. Then
\[
 b_t=(1-A_t)p_{01,t}-A_t p_{10,t},\qquad
 (b_t^-)^2=A_t p_{10,t}^2,\qquad
 b_t^2=A_t p_{10,t}^2+(1-A_t)p_{01,t}^2.
\]
\end{proposition}
\begin{proof}
Condition on $A_t=0$ or $A_t=1$ in
$b_t=\E[A_{t+1}-A_t\mid\F_t]$. Only the latter case permits negative
drift; squaring gives the two identities.
\end{proof}

Thus the one-sided objective penalizes squared loss hazard at correct
states, whereas the symmetric objective also penalizes repair at incorrect
states. A binary martingale is constant almost surely; a submartingale can
repair errors while making the correct set absorbing. These statements do
not supply an observable correctness oracle or a tested training method.

\begin{corollary}[Adverse energy bounds stopping gain]
\label{cor:energygain}
Under Cor.~\ref{cor:advantage}, define
$\mathcal E_-=\E[\sum_{t<\tau}(b_t^-)^2]$. For any admissible $S\le\tau$,
\[
 \E[C_S-Y]\le\min\{1,\sqrt{\E[\tau]\mathcal E_-}\}.
\]
\end{corollary}
\begin{proof}
The stopping identity bounds the gain by
$\E[\sum_{t<\tau}b_t^-]$. Cauchy--Schwarz on the product measure of
paths and pre-termination time indices bounds this by
$\sqrt{\E[\tau]\E[\sum_{t<\tau}(b_t^-)^2]}$. Its total measure is
$\E[\tau]<\infty$; $|b_t|\le1$ ensures integrability. The gain is also
at most one.
\end{proof}
For binary commitments, $\E[\sum_{t<\tau}b_t^-]$ equals the expected
number of correct-to-wrong transitions. The reported early-correct/final-wrong
frequency is not this count: it omits later onset and losses subsequently
repaired, so it cannot be inserted into this bound.

\section{Protocol-Matched Continuation Follow-Up}
\label{app:matched}

\paragraph{Design and selection.}
This post-hoc follow-up was specified in an internal plan after the primary
results were known, with the sample fixed before collecting the new outcomes.
All 12 historical wrong-ending 1.5B chains with a correct nonterminal
checkpoint at or before 2048 are included, across all five MATH levels.
For each first-correct checkpoint, one unique right-ending chain with a
correct answer at that same checkpoint is selected, with an exact difficulty
match. A seeded shuffle followed by minimum level-distance matching gives
the 12 pairs in Table~\ref{tab:matchedpairs}; no new outcome enters selection.
All 24 immediate greedy answers reproduce the archived correct label.

\paragraph{Protocol fidelity and runtime.}
The cached R1-Distill-Qwen-1.5B model is run in FP16 on the same 12\,GB
RTX~4070~Ti. Prompt tokens are reconstructed with the original chat template;
continuations start from the original generated prefix token IDs, without
decoding and retokenizing that prefix. At checkpoint $t$, each of eight
continuations has $4096-t$ new-token allowance, temperature $0.6$, and top-$p$
$0.95$, ending at EOS or the cap. Batches contain four draws; a deterministic
per-prefix/per-batch seed derived from 20260907 permits exact resumption within
the recorded runtime. The frozen manifest records input and source hashes,
model revision, sampling configuration, and selected IDs. Runtime is
PyTorch~2.13.0+cu126 and Transformers~5.15.1; cross-version bitwise reproduction
of historical generation is not assumed.

Immediate forcing and terminal forcing reproduce the original text-based
operator: close the thought if necessary, append the boxed-answer cue, and
decode greedily for at most 48 tokens. At termination, a boxed answer after
the closing thought is used if present; otherwise the forcing operator is
applied. Forced outputs use the original brace-closing parser and symbolic
verifier. A strictly balanced-box parser is also evaluated and changes no
correctness label among the 192 continuations. No continuation lacks a scored
answer. Terminal forcing is used in $85/96$ case and $28/96$ control draws;
respectively $11/96$ and $68/96$ reach EOS. Thus the fixed-budget protocol
still involves substantial censoring, but unlike the historical branching
diagnostic it does not score a missing unforced answer as an automatic error.
Raw tokens, answer strings, scoring flags, prompt/generation token counts,
and timings are retained. Generation of the continuations and their terminal
forcing uses 405,150 tokens including EOS, excluding the immediate preflight
checks; this is not a matched FLOP or wall-clock policy comparison.

\paragraph{Estimates and uncertainty.}
For arm $a\in\{\mathrm{case},\mathrm{control}\}$ and pair $i=1,\ldots,12$,
let $Y_{aij}$ be terminal correctness of continuation $j=1,\ldots,8$.
The fixed immediate answer is correct at every selected prefix, so
\[
 \widehat\Phi_{ai}=\frac18\sum_{j=1}^8Y_{aij},\qquad
 \widehat d_{ai}=1-\widehat\Phi_{ai},\qquad
 \widehat D=\frac1{12}\sum_{i=1}^{12}
 (\widehat d_{\mathrm{case},i}-\widehat d_{\mathrm{control},i}).
\]
Thus prefixes have equal weight; the 192 draws are not 192 independent
problems. The contrast is descriptive of these selected cohorts, not a
randomized treatment effect.
Each arm has 12 prefixes and 96 new continuations. Immediate correctness is
one at every prefix; continuation success is $\matchedCaseCorrect/96$ in cases
and $\matchedControlCorrect/96$ in controls. The equal-prefix mean gaps are
$\matchedCaseGap$ (95\% prefix-bootstrap interval $\matchedCaseGapCI$) and
$\matchedControlGap$ ($\matchedControlGapCI$) percentage points.
The case-minus-control contrast is $\matchedGapContrast$ points
($\matchedGapContrastCI$). The 5000-resample paired bootstrap draws entire
matched pairs and recomputes the two means; it summarizes pair-level sampling
variation rather than providing an exact population interval. Per-prefix
Wilson intervals in Table~\ref{tab:matchedpairs} expose the limited precision
of eight Monte Carlo draws. Conditional plug-in Monte Carlo standard errors
of the two arm means are 3.43 and 1.36 percentage points, but can understate
uncertainty at all-success/all-failure prefixes; the Wilson intervals should
not be replaced by zero uncertainty there.

Specifically, NumPy's default random generator with seed 20260907 draws
5000 vectors of 12 pair indices with replacement. The same indices are used
for both arms, and the reported limits are the 2.5th and 97.5th percentiles.
Draws are not independently resampled inside prefixes. These descriptive
intervals therefore do not separately propagate all within-prefix Monte
Carlo uncertainty or confer coverage for an outcome-independent population.
For an arm with success estimates $\widehat\Phi_i$, the reported conditional
plug-in Monte Carlo standard error is
$\sqrt{\sum_{i=1}^{12}\widehat\Phi_i(1-\widehat\Phi_i)/7}/12$.
The per-prefix Wilson intervals use eight Bernoulli trials and
$z=\Phi_{\mathrm N}^{-1}(0.975)$, with $\Phi_{\mathrm N}$ denoting the
standard normal distribution function. These summaries quantify different
sources of variation and are not interchangeable.

\begin{table}[t]
\centering\small
\caption{All matched pairs. IDs index MATH-500; the first ID is the historical
wrong-ending case and the second its right-ending control. Every immediate
answer is correct. Brackets are per-prefix 95\% Wilson intervals for
continuation success, not intervals for population prevalence.}
\label{tab:matchedpairs}
\begin{tabular}{@{}rrrcccc@{}}
\toprule
Case/control & Level & Checkpoint & Case correct & Interval & Control correct & Interval\\
\midrule
\matchedPairRows
\bottomrule
\end{tabular}
\end{table}

\paragraph{Scope of the inference.}
At a specified prefix, the matched success gap estimates expected remaining
adverse drift by Eq.~\eqref{eq:gap}. It does not identify the mechanism of a
particular flip. Because the cohorts were selected using archived final
outcomes, this follow-up does not revise the $\rEightRate\%$ prevalence,
estimate loss at an outcome-independent sample of prefixes, or validate a
deployable stopping policy. Difficulty and checkpoint matching do not remove
all differences between problems. The historical cross-protocol diagnostic
in App.~\ref{app:expdetails} is preserved separately; its mean continuation
success is not pooled with this experiment.

\section{Whole-Problem Refit Sensitivity}
\label{app:refits}

This post-hoc diagnostic tests sensitivity to the fitted measurement pipeline,
not a new behavioral intervention. Each model uses its original layer-21
states and problem-disjoint outer probe/critic/evaluation split. Logistic
regularization is fixed at $C=0.1$, rather than retuned by the original
cross-validation grid. Within the probe role, the original 20\% calibration
problems remain disjoint from probe fitting. A fixed 15\% of eligible
critic-role problems form a disjoint validation set, replacing the original
step-level validation partition. Ten seeds, 20260907--20260916, independently
resample whole problems with replacement within each development role and
final-label stratum, then refit both probes and both critics. The evaluation
problems never enter fitting, calibration, early stopping, or model selection.

Probe fitting uses logistic regression (maximum 2000 iterations) and
temperature calibration. The critic retains the two-layer 256-unit MLP,
dropout $0.1$, AdamW with learning rate $10^{-3}$ and weight decay $10^{-4}$,
batch size 1024, maximum 200 epochs, and validation-NLL early stopping with
patience 12. Dense paragraph increments and normalized token position follow
the primary pipeline. Every fitted pipeline is evaluated on the same held-out
problems using $\widehat V/\widehat{\mathrm{Var}}(\Delta)$; the two readouts
use identical evaluation steps. Complete development resampling lists, probe
AUC/ECE/Brier scores, temperatures, critic validation diagnostics, and
per-step held-out predictions and targets are saved.

\begin{table}[ht]
\centering\small
\caption{All ten refits of the diagnostic variant. C/V denote commitment/verdict
scores $\widehat V/\widehat{\mathrm{Var}}(\Delta)$; gap is C minus V. Ranges
across these refits are sensitivity summaries, not confidence intervals.}
\label{tab:refits}
\begin{tabular}{@{}r rrr rrr@{}}
\toprule
 & \multicolumn{3}{c}{1.5B} & \multicolumn{3}{c}{7B}\\
\cmidrule(lr){2-4}\cmidrule(l){5-7}
Refit & C & V & Gap & C & V & Gap\\
\midrule
\refitDetailRows
\bottomrule
\end{tabular}
\end{table}

Commitment exceeds verdict in $\refitCasePositive/10$ and
$\refitControlPositive/10$ refits, respectively. The mean gaps are
$\refitSmallGapMean$ and $\refitLargeGapMean$; Table~\ref{tab:refits} retains
all fits, including the negative 1.5B commitment score in refit 2. That probe
has temperature 101.02, ECE 0.020, and increment variance
$2.22\times10^{-4}$, while its critic predicts mean increment 0.0137 against
observed mean 0.00017 and performs worse than the zero predictor in the
variational score. This is an observed sensitivity, not a reason to discard
the run or evidence for a unique cause. A small ECE does not guarantee drift
estimation quality. With ten development refits and fixed evaluation problems,
these results are not a reliable 95\% interval for full-pipeline uncertainty.
They qualify the original single-fit ordering without replacing its numerical
results or resolving the real-model verdict null.

\section{Supplement Guide and Reproduction Boundaries}
\label{app:artifacts}

The unified \texttt{supplementary\_code.zip} contains the primary audit and
the separately versioned follow-up, together with archived manuscript sources
and audit figures. App.~\ref{app:beyond} adds no experimental results. The root README gives an artifact index and commands.
The original result files and numerical macros are retained; new measurements
are isolated under \texttt{results/research\_20260907/} and
\texttt{paper/followup\_numbers.tex}. Internal plans record when designs were
fixed, not external registration or independent replication.

\begin{center}
\small
\begin{tabular}{@{}p{0.26\textwidth}p{0.69\textwidth}@{}}
\toprule
Claim or appendix & Evidence and verification route\\
\midrule
Primary audits; App.~\ref{app:expdetails} &
Original \texttt{results/}, \texttt{results7b/}, \texttt{results\_qwen3/},
\texttt{results\_llama8b/} and \texttt{results\_phi4mini/} JSONs;
\texttt{code/make\_numbers.py} regenerates the historical macro file.\\
Observation control; App.~\ref{app:observation} &
\texttt{code/research\_exact\_control.py} enumerates the four paths with
rational arithmetic; \texttt{exact\_control.json} stores the 29 checks.\\
Matched continuations; App.~\ref{app:matched} &
\texttt{matched\_manifest.json}, \texttt{matched\_runtime.json},
\texttt{matched\_immediate.jsonl}, \texttt{matched\_draws.jsonl}, and
\texttt{matched\_summary.json}; all are in the follow-up results directory.\\
Refits; App.~\ref{app:refits} &
\texttt{refit\_config.json}, role lists, 20 individual JSON/NPZ output pairs,
and \texttt{refit\_summary.json}; NPZ files contain held-out targets,
critic predictions, and problem IDs.\\
\bottomrule
\end{tabular}
\end{center}

\paragraph{Model-free verification.}
The archive provides a standard-library SHA-256 integrity check. From a
fresh extraction, run:
\begin{quote}\small\ttfamily
python code/verify\_supplement.py
\end{quote}
Adding
\verb|--evidence| requires NumPy and recomputes the matched success rates,
paired-bootstrap intervals, per-prefix Wilson intervals, token totals,
parser-label agreement, and all saved refit scores. It also checks frozen
source hashes, problem-role separation, numerical follow-up macros, and the
exact finite-state construction. Checks run in a temporary copy and leave
the packaged evidence unchanged. They verify arithmetic and consistency,
not the validity of the original answer verifier or independent replication
of model behavior.

\paragraph{Additional assets and computation requirements.}
The archived 1.5B rollout and forced-answer files needed for the matched
manifest are included. Model weights, dataset caches, and the large primary
hidden-state arrays are not. New continuations require the recorded model
revision, dataset, generation runtime, and GPU resources. Repeating the
refits requires the original layer-21 state arrays and index files for both
models; saved held-out predictions suffice only for score recomputation,
not refitting. The historical README documents the extraction pipeline.
Different runtime versions can change generated trajectories and therefore
all downstream fitted quantities. Independent reruns must use a new
output location or version, preserving the frozen evidence.

\paragraph{Evidence that remains absent.}
The matched experiment does not sample prefixes independently of archived
outcomes, and the refit diagnostic is neither an original-tuning replication
nor a full-pipeline confidence interval. Neither experiment supplies a
held-out deployment benefit or identifies the cause of the verdict residual.
The historical cross-protocol branching study and randomized-refit ranks
remain archived diagnostics with their original limitations; they are not
silently replaced by the new follow-up.

\section{Beyond Single-Chain Reasoning: Scope and Testable Extensions}
\label{app:beyond}

The central distinction is between forecasting a fixed protocol's eventual
success and evaluating the artifact that could be delivered now. The unit
of computation need not be a token or a single-model reasoning segment.
This section makes that extension explicit, without claiming empirical
validation outside the settings in Sec.~\ref{sec:experiments}.

\paragraph{A protocol-level formulation.}
Fix a finite horizon $T$ and a complete stochastic protocol, including its
history-dependent choices of models, tools, messages and updates. Let
$\mathcal H_t$ be its nested full-history filtration, $C_t\in[0,1]$ the
score obtained by a specified deliver-now operator, and
$u_t=\E[C_t\mid\mathcal H_t]$ its current commitment. Fix the scoring rule
and evaluation distribution. Let $Y\in[0,1]$ be the terminal score and
require a common terminal boundary $u_T=Y$, with the terminal evaluation
included in $\mathcal H_T$. Define
\[
 v_t=\E[Y\mid\mathcal H_t],\qquad
 d_t=\E[u_{t+1}-u_t\mid\mathcal H_t].
\]
The tower property and conditional telescoping give
\begin{equation}
 \E[v_{t+1}\mid\mathcal H_t]=v_t,\qquad
 u_t-v_t=-\E\!\left[\sum_{s=t}^{T-1}d_s\,\middle|\,\mathcal H_t\right].
 \label{eq:protocolgap}
\end{equation}
For a history-measurable stopping time $S\le T$, the same argument yields
$\E[C_S-Y]=-\E[\sum_{s=S}^{T-1}d_s]$. Early termination can be represented
by freezing the score thereafter. These are applications of the existing
identities to a bounded score and a joint protocol, not new mechanisms or
guarantees of improvement. They need no independence among agents or
Markov assumption; a state-based Poisson equation additionally requires a
sufficient Markov state. If the stopping readout and terminal score do not
satisfy the common-boundary condition, a terminal
mismatch term must be retained.

\paragraph{Multi-step reasoning and iterative self-revision.}
For a single task, let the current artifact be the answer after each
reasoning segment or completed critique--revision round. Self-Refine, for
example, iteratively generates feedback and revises an output
\citep{madaan2023selfrefine}. The commitment is the current artifact's
externally scored quality; the verdict forecasts the final artifact under
the remaining refinement protocol. More elaborate feedback, increased
confidence, or a changed answer does not alone identify their difference.
A direct audit samples prefixes without selecting on their eventual
outcomes, scores the immediate artifacts, and branches fresh continuations
under the same protocol. This estimates remaining improvement or erosion
at those states, while repeated runs quantify sampling variability.
The theory permits beneficial revision and does not prescribe zero drift:
positive drift is precisely the improvement that refinement seeks.

\paragraph{Multi-agent collaboration and debate.}
Multiagent debate uses several model instances to exchange and revise
reasoning before producing a common answer \citep{du2023multiagent}.
Here $\mathcal H_t$ contains the joint interaction history, including
relevant private states and tool observations, and the deliver-now operator
specifies how the current messages or candidate answers become one system
output. Equation~\eqref{eq:protocolgap} concerns that system output, not
the average confidence or agreement of its members. A transcript-only
probe may be lossy; even an exact forecast conditioned on one agent's
local information need not be a martingale relative to the larger joint
filtration. Thus the information set must be specified before interpreting
predictable updates as a control failure.

Communication can change both dependence and expected answer quality.
The independent-ensemble $1/\sqrt{k}$ innovation scaling discussed in
App.~\ref{app:proofs} therefore does not automatically extend to interacting
agents. Neither consensus nor correlated errors alone determine the sign of
remaining drift. A useful audit fixes the final-answer aggregation rule,
measures quality and agreement separately at each round, and resamples
continuations from the same joint state. Comparisons among communication
protocols require their own verdicts and compute accounting. If only
unfinished problem instances remain in a later-round average, the survivor
decomposition applies to that changing problem cohort; replacing an agent
inside a fixed problem is not itself this population-selection effect.

\paragraph{Recursive improvement of the system itself.}
Improving an answer is distinct from modifying the machinery that generates
answers. The Darwin G\"odel Machine, for example, modifies agent code and
evaluates the resulting agents on coding benchmarks \citep{zhang2025dgm}.
To formulate an analogous audit, the state must include the evolving code,
parameters, memory, selection history and update procedure. The current
artifact is now the system version that would be deployed if development
stopped, and commitment measures its performance under a fixed evaluation
distribution. A fixed \emph{meta-protocol} may contain adaptive, even
self-modifying, updates: ``fixed'' specifies the overall probability law,
not frozen model parameters.

Under that specification and the common terminal boundary,
Eq.~\eqref{eq:protocolgap} still concerns the forecast of one terminal
evaluation. It neither fixes the performance of intermediate versions nor
rules out recursive capability improvement. Externally changing the update
procedure or evaluation target changes the comparison; a verdict for the
old protocol cannot simply be carried over. A frozen probe also need not
retain its meaning when the underlying representations change
(App.~\ref{app:deploy}). A credible empirical extension would separately
evaluate successive deployable versions on held-out tasks, account for
selection and development compute, and use replicated update trajectories
where feasible. Our present experiments establish none of these
system-improvement claims.

\paragraph{Implication for LLM development.}
Across these settings, the framework supplies a common comparison target:
the externally evaluated artifact available now versus the artifact
expected after additional computation. It separates measurement from
mechanism and decision quality. This can inform how future evaluations,
process supervision and adaptive-compute methods are designed, but it does
not establish that our present estimators or stopping heuristics transfer
successfully to these systems.

\section{Probe Diagnostics, Synthetic Validation, and Stopping Replays}
\label{app:diagnostics}

This appendix collects the learned-sensor and policy experiments.
The main-text behavioral results use scored answers directly; the analyses
here evaluate hidden-state measurement and cached-checkpoint control.

\subsection{Readout estimation}
\paragraph{Probe estimation and observation loss.}
Identification specifies the exact quantity; estimation must recover it.
For a lossy tap $z_t=f(h_t)$, a critic observes only
$m_t=\E[\Delta_t\mid z_t,t]$; conditional Jensen gives
$\E[m_t^2]\le\E[b_t^2]$. This population bound concerns the \emph{same}
readout, not an arbitrary learned substitute. If $\hat\pi_t=\pi_t+e_t$, its
full-filtration drift differs by $\E[e_{t+1}\mid\F_t]-e_t$, at most
$2\epsilon$ when $|e_t|\le\epsilon$ uniformly
(Prop.~\ref{prop:readouterror}). AUC and binned calibration error do not
establish that bound.

The commitment probe predicts current forced-answer correctness; the verdict
probe predicts terminal correctness. An exact verdict conditioned on a nested
observation history is a martingale, even without full-state sufficiency.
Current-state taps need not retain that history: even
$v_t=\E[Y\mid z_t,t]$ can have predictable increments
(App.~\ref{app:observation}). Thus calibration alone does not establish the
verdict null. Our control remains unresolved (Sec.~\ref{sec:validity});
matched behavioral continuations provide a probe-independent check on loss.

\subsection{Variational objectives and critic implementation}
\paragraph{Training implication: stability is not improvement.}\label{sec:objective}
The decomposition also distinguishes which changes a regularizer suppresses.
A quadratic increment penalty charges both predictable change and innovation:
$\E[\Delta_t^2]=\E[b_t^2]+\E[\eta_t^2]$ (Prop.~\ref{prop:quadratic}). Because
$m_t=\E[\Delta_t\mid z_t,t]$ is an $L^2$ projection,
$\E[m_t^2]=\max_g\{2\E[g(z_t,t)\Delta_t]-\E[g(z_t,t)^2]\}$ with $g^\star=m_t$,
and penalizing this variational value---requiring no new labels given the readout, blind to innovation
variance---vanishes iff the chain is \emph{state-calibrated} ($m_t\equiv0$).
This symmetric objective penalizes beneficial predictable improvement too.
For erosion alone, a sign-constrained variant targets
$\E[(m_t^-)^2]=\sup_{g\le0}\E[2g\Delta_t-g^2]$, where
$m_t^-=\max\{-m_t,0\}$ (App.~\ref{app:objective}); we derive, but do not
experimentally evaluate, this variant.
Suppressing movement, drift, and erosion are different objectives
(Apps.~\ref{app:objective}--\ref{app:deploy}); only synthetic validation is
provided (Sec.~\ref{sec:exp0}).

\paragraph{The critic as a sensor.}\label{sec:control}
Regress frozen-model probe increments on $(z_t,t)$ and accumulate
$\hat B_t=\sum_{s<t}g_\phi(z_s,s)$. With a sufficient tap and uniform critic
error $\delta$, compensating the exact readout by $\tilde\pi_t=\pi_t-\hat B_t$
gives $|\E[\tilde\pi_\tau-\tilde\pi_0]|\le\delta\E[\tau]$
(Prop.~\ref{prop:graceful}, App.~\ref{app:proofs}). This is a readout-error
bound, not a guarantee of improved answers or of accurate learned probes.
Our diagnostic \emph{drift alarm} stops when $-\hat B_t>\kappa$, using the
last checkpoint answer; $\kappa$ is pre-registered as the 90th percentile
of $\max_t(-\hat B_t)$ over \sco{} critic-split chains. Accumulated past drift
is a heuristic signal, not the remaining-drift oracle above.

\paragraph{Minimal audit recipe.}
(i)~Elicit answers at fixed intermediate budgets.
(ii)~Measure correct-to-wrong prevalence with repeated samples within problems.
(iii)~Separate within-cohort and composition changes.
(iv)~Test a sensor or stopping rule where behavioral checks show expected
loss at candidate states; aggregate deterioration is not required.
The first three steps assess measurement; the fourth assesses control
(App.~\ref{app:deploy}).

\subsection{Experiment 0: apparatus calibration with known injected drift}
\label{sec:exp0}

A binary Gaussian-evidence process supplies known drift: its unperturbed
posterior on a fixed hypothesis is a martingale, while a constant log-odds
lean injects probability-scale drift. With evidence separation $0.35$, noise
SD $1$, and $60$ steps, the critic recovers it (slope $\eZeroSlopeMid$,
$R^2=\eZeroRTwoMid$ at $\varepsilon=0.05$), reports no false drift at zero,
and has matched-sample detection floor $\eZeroFloorMatched$.
Under decaying evidence, accuracy loses $\eZeroOverthinkDelta$ points after
peaking at $T^\dagger=\eZeroPeakT$; compensation restores it to
$\eZeroCompFinal\%$. The variance-blind penalty preserves gain and accuracy
over the tested $\lambda$ values, whereas $\E[\Delta\pi^2]$ collapses both.
This calibrates the apparatus, not real-model decision accuracy; the
19-configuration sweep and late-onset test are in App.~\ref{app:expdetails}.

\subsection{Primary probe measurements}
\paragraph{Predictable probe increments, with positive mean predicted drift.}
Here $\Delta=\Delta\hat\pi_t$ and $m=\E[\Delta\mid z_t,t]$ refer to the
\emph{probe} increments. For any fixed critic $g$, the population value
$V(g)=\E[2g\Delta-g^2]=\E[m^2]-\E[(g-m)^2]\le\E[m^2]$ is a lower bound
on state-conditional drift energy. Its held-out sample estimate $\widehat V$
is not itself a guaranteed finite-sample bound. We report the normalized score
$\widehat V/\widehat{\mathrm{Var}}(\Delta)$: $\vCommit$ $\vCommitCI$ at 1.5B
and $\sevBvCommit$ $\sevBvCommitCI$ at 7B, with chain-bootstrap intervals
conditional on the fitted probes and critics. This variance normalization is
not a bounded fraction of increment energy. Mean predicted increments are
positive in all four label groups, with \sco{} CIs excluding zero
($\gdSCci$/$\sevBgdSCci$). The critic also predicts positive mean drift for
\ow{} chains and cannot resolve their rare erosion. Evidence for aggregate
net improvement comes from the behavioral within-cohort term
(Sec.~\ref{sec:exp1b}); it does not exclude adverse drift in particular states.
Randomized-refit diagnostics and their inferential limitations are in
App.~\ref{app:audit2}.

\begin{table}[t]
\centering\footnotesize
\caption{Primary probe scores. Brackets are conditional chain-bootstrap
95\% intervals. These scores characterize learned readout increments.}
\label{tab:probesummary}
\begin{tabular}{@{}lcc@{}}
\toprule
 & R1-Qwen 1.5B & R1-Qwen 7B\\
\midrule
Commitment score & $\vCommit$ $\vCommitCI$ & $\sevBvCommit$ $\sevBvCommitCI$\\
Verdict score & $\vVerdict$ & $\sevBvVerdict$\\
Commitment minus verdict & $\vGap$ $\vGapCI$ & $\sevBvGap$ $\sevBvGapCI$\\
\bottomrule
\end{tabular}
\end{table}

\subsection{Held-out stopping-policy replays}
\label{sec:exp2}

We fix thresholds on the critic split and replay policies using only
information available at each decision, on held-out evaluation chains
($n=110/80$ at 1.5B/7B). Table~\ref{tab:stopping} compares the drift alarm,
two-consecutive-answer agreement, checkpoint voting, and full budget.
Charged tokens are generated length plus 48 per forcing branch, with no
terminal forcing charged to the archived full-budget baseline; prefill and
critic costs are excluded. This is not a wall-clock or FLOP comparison.

None of the implementable configurations ($\blNimpl/\blNmethods$ at 1.5B;
$\sevBblNimpl/\sevBblNmethods$ at 7B), or alarm thresholds from the 50th to
99th healthy-chain percentile, improves accuracy at lower charged token
cost. Token-saving rules lose $\blSaveMild$ to $\blSaveWorst$ points at
1.5B and $\sevBblSaveMild$ to $\sevBblSaveWorst$ at 7B; only
$\blSaveCIzero/\blSaveN$ and $\sevBblSaveCIzero/\sevBblSaveN$ intervals
reach zero. The alarm fires on $\alarmFireSC\%/\sevBalarmFireSC\%$ of
\sco{} chains and catches none/one of the two evaluation \ow{} chains.
App.~\ref{app:audit2} documents native-method deviations and the full replay
matrix. These results assess our cached-checkpoint protocol, not the
cited methods under their native execution.

The negative result distinguishes a quantity's theoretical relevance from
our ability to estimate and act on it. It establishes no general
impossibility of beneficial early exit.

\begin{table}[t]
\centering\footnotesize
\caption{Held-out stopping replay: accuracy (\%) and mean charged tokens, on
110/80 evaluation chains at 1.5B/7B with only two early-correct/final-wrong
chains each. Small differences do not establish a policy ranking;
App.~\ref{app:audit2} gives the full $\blNmethods$-configuration gauntlet.}
\label{tab:stopping}
\begin{tabular}{@{}lrrrr@{}}
\toprule
 & \multicolumn{2}{c}{R1-Distill-Qwen 1.5B} & \multicolumn{2}{c}{R1-Distill-Qwen 7B}\\
\cmidrule(lr){2-3}\cmidrule(lr){4-5}
Policy & Accuracy & Tokens & Accuracy & Tokens\\
\midrule
Full budget & $\fullAcc$ & $\fullTokens$ & $\sevBfullAcc$ & $\sevBfullTokens$\\
Drift alarm & $\alarmAcc$ & $\alarmTokens$ & $\sevBalarmAcc$ & $\sevBalarmTokens$\\
Checkpoint vote & $\voteAcc$ & $\voteTokens$ & $\sevBvoteAcc$ & $\sevBvoteTokens$\\
Answer agreement & $\agreeAcc$ & $\agreeTokens$ & $\sevBagreeAcc$ & $\sevBagreeTokens$\\
\bottomrule
\end{tabular}
\end{table}
\subsection{Validity: the unresolved verdict control, and robustness}
\label{sec:validity}
\paragraph{The tower-property control remains unresolved.}
Score ordering changes across scales and taps (Table~\ref{tab:probesummary}).
Ten whole-problem bootstrap refits per model, using disjoint fitting,
calibration and validation problems and a fixed evaluation split, give
commitment above verdict in only $\refitCasePositive/10$ and
$\refitControlPositive/10$ runs. This is tuning sensitivity, not a confidence
interval (App.~\ref{app:refits}). Verdict AUCs
($\verdictAUCthree$/$\sevBverdictAUCthree$) also miss the prespecified $0.75$
bar. An exact verdict on nonnested observations can drift
(App.~\ref{app:observation}); adding a cumulative history summary leaves
negative gaps ($\ntNestedGap$/$\sevBntNestedGap$), but does not guarantee
nested information. Neither test separates observation loss from estimation
error (App.~\ref{app:audit2}).

\paragraph{Sensitivity of the behavioral findings.}
Absent the forcing string, $\fvNoAnsPct\%$ of $\fvPairs$ truncated prefixes emit
no answer at all---the reason an operator is needed; where an unforced
continuation does answer, it agrees with the forced answer on correctness in
$\fvCorrAgree\%$ of cases, and greedy forcing exceeds sampled by $\dmDelta$
points ($\dmCI$). The \ow{} count is grid-dependent
($\gCoarseOW$/$\gMidOW$/$\gDenseOW$ at 1024/512/256 tokens, matched
$n=\gDenseN$), so ${\approx}3\%$ is specific to the 512-token grid. Prevalence
stays at 1--4\% under every final-outcome definition and at temperature $0.9$, the
within-cohort term stays positive, and a linear critic retains the drift signal;
on AIME-2024 ($\aimeChains$ chains) no chain is \ow{}, though only $\aimeElig$
are ever correct early (App.~\ref{app:expdetails}).
Table~\ref{tab:auditfull} and App.~\ref{app:audit2} give each check with its
interval.

\section{Analyses Added During Review}
\label{app:rebuttal}

\paragraph{When each analysis was specified.} Table~\ref{tab:status} lists every
analysis by the time its design was fixed. ``Before results'' means written in
a dated internal plan before any real-data result was computed; the plan is
internal, not an external registry. All three categories fixed their design
before generating their own data, and every outcome is reported.

\begin{table}[h]
\centering\footnotesize
\caption{\textbf{When each analysis was specified.}}
\label{tab:status}
\begin{tabular}{@{}p{0.47\textwidth}p{0.25\textwidth}p{0.2\textwidth}@{}}
\toprule
Analysis & Specified & Reported in\\
\midrule
Early-correct/final-wrong prevalence (Gate 1a), 7B escalation, predictable
commitment drift, stopping-rule replays & before results & Secs.~\ref{sec:exp1},
\ref{sec:stopping}; App.~\ref{app:expdetails}\\
Eight-sample resampling, temperature 0.9, 256-token grid, greedy vs.\ sampled
forcing, Qwen3-1.7B, self-consistency & follow-up, before its run &
Sec.~\ref{sec:exp1b}; App.~\ref{app:audit2}\\
R1-Distill-Llama-8B and Phi-4-mini-reasoning audits & follow-up, before its run &
Table~\ref{tab:audit}\\
Survivorship decomposition; matched-prefix continuations; probe refits;
cumulative-history ablation & follow-up after primary results; continuation and
refit designs fixed before their runs & Secs.~\ref{sec:exp1}--\ref{sec:exp1b};
Apps.~\ref{app:matched}, \ref{app:refits}\\
Outcome-blind continuations, generic doubt, misleading hint, 16k budget,
1.5B in NF4, 7B in 8-bit, OlympiadBench, Qwen3.5-2B, problem concentration,
difficulty match, oracle headroom &
during review, before its run & Secs.~\ref{sec:exp1}--\ref{sec:stopping};
this appendix\\
\bottomrule
\end{tabular}
\end{table}

\paragraph{Outcome-blind continuations.} The state set contains every primary
1.5B chain with a correct non-terminal forced answer at or before 2048 tokens,
at its first such checkpoint: $\cenN$ states across all five MATH levels
($\cenNlvl$ at levels 3--5). The rule reads no final outcomes; the 12 states of
the matched study (App.~\ref{app:matched}) are among them. The protocol copies
the matched study: original prefix token ids, temperature $0.6$, top-$p$
$0.95$, the remaining budget $4096-t$, the post-think boxed answer if present
and otherwise a greedy forced answer of at most 48 tokens. Four continuations
were drawn per state. Before any continuation, the immediate greedy answer
reproduced the archived correct answer at every state. Intervals resample
problems. Continuations end correct in $\cenCont\%$ of draws; the mean loss is
$\cenLoss$ points ($\cenLossCI$). By state, $\cenKeepAll$ keep the answer in
all four draws, $\cenAnyLoss$ lose it at least once, $\cenHalf$ in at least
two, and $\cenZero$ in all four. The strict balanced-box parser gives the same
scores.

\paragraph{Generic doubt and misleading hint.} Both manipulations use the same
$\injN$ states, a seeded random subset of the outcome-blind set, and are paired
with the outcome-blind draws at those states. The generic doubt appends ``Wait,
I made a mistake in an earlier step, so this answer is incorrect. Let me redo
the solution carefully.'' A forced answer right after it is correct in
$\injImm\%$ of states; continuations end correct in $\injCont\%$ against
$\injCtrl\%$ without it (paired difference $\injDrop$ points, $\injDropCI$).
The misleading hint appends ``Wait, I made an error earlier. Rechecking
carefully, the correct answer should be~$W$.'' The wrong answer $W$ is the most
frequent wrong answer the 1.5B model gave on that problem in our runs
($\hintModel$ states), otherwise the gold answer plus one for integer answers
($\hintPlus$ states); $\hintExcl$ states have neither and are excluded. A forced
answer right after the hint is correct in $\hintImm\%$ of states and equals $W$
in $\hintImmAdopt\%$. Continuations end correct in $\hintCont\%$ against
$\hintCtrl\%$ without the hint, and end at $W$ in $\hintAdopt\%$.

\paragraph{Problem concentration.} With eight fresh chains per problem (367
problems), the number of early-correct/final-wrong chains per problem has
$\propOD$ times the variance expected if all problems shared one rate
(permutation test against that null, $p<10^{-4}$). The intraclass correlation
is $\propICC$: about a quarter of the variation lies between problems. The
transitions occur in $\propAny$ problems, against $\propAnyNull$ expected under
a common rate, and the ten most affected problems produce $\propTopTen\%$ of
them. Restricting to problems whose chains ever give a correct early answer
leaves these figures essentially unchanged.

\paragraph{Difficulty and precision.} On the 262 level-4--5 problems used for
the 7B model, the 1.5B FP16 audit gives $\dfmOW/\dfmN=\dfmPct\%$ ($\dfmCI$)
early-correct/final-wrong chains, the same count as the 7B model, with final
accuracy $\dfmAcc\%$. Rerunning the 1.5B model in NF4 on the same problems
gives $\nfOW/\nfN=\nfPct\%$ ($\nfCI$) and final accuracy $\nfAcc\%$. The 7B
model loaded in 8-bit (bitsandbytes LLM.int8) on its 262 problems, seed 0,
gives $\ieOW/\ieN=\iePct\%$ ($\ieCI$) with final accuracy $\ieAcc\%$; its
decomposition is in Table~\ref{tab:audit}.

\paragraph{Second benchmark and a 2026 model.} OlympiadBench: from the
text-only, English, open-ended competition-mathematics split, we kept problems
with a single numerical or expression answer and no unit (558), and drew a
seeded sample of 367 to match the MATH-500 primary set. The 1.5B FP16 model ran
with the primary protocol (seed 20261108). It reaches final accuracy
$\obAcc\%$, with $\obOW/\obN=\obPct\%$ ($\obCI$) early-correct/final-wrong
chains. Qwen3.5-2B ran in FP16 on the 367 level-3--5 MATH-500 problems with its
chat template's thinking mode enabled and otherwise the primary protocol
(seed 0). It reaches final accuracy $\qfAcc\%$, with $\qfOW/\qfN=\qfPct\%$
($\qfCI$). Decompositions are in Table~\ref{tab:audit}.

\paragraph{Longer budget.} Fresh 1.5B chains (seed 20261107) on the 367
level-3--5 problems ran to a 16{,}384-token cap with forced answers every 1024
tokens. Because generation is left to right, forced answers at checkpoints up
to 4096 tokens are unaffected by the cap, so the same chains also give the
4096-token audit; a chain still running at 4096 tokens counts as capped there.
At the two caps, $\lbShortCapped\%$ and $\lbLongCapped\%$ of chains are capped,
final accuracy is $\lbShortAcc\%$ and $\lbLongAcc\%$, and the
early-correct/final-wrong rate is $\lbShortPct\%$ ($\lbShortCI$) and
$\lbLongPct\%$ ($\lbLongCI$). The coarser grid finds no more transitions than
the 512-token grid (Sec.~\ref{sec:exp1b}), so these rates are comparable with
each other rather than with the primary rate. Over 4096 to 16{,}384 tokens the
decomposition gives within-cohort $\lbWithin$ ($\lbWithinCI$) and composition
$\lbComp$ points.

\paragraph{Oracle headroom.} The oracle knows the gold answer and stops each
held-out chain at its cheapest correct checkpoint, otherwise running to the
full budget. It is an upper bound for any rule that stops on the information in
a chain, under the same answer procedure and token accounting.

\end{document}